\documentclass[10pt,twocolumn,letterpaper]{article}

\usepackage{iccp}
\usepackage{times}
\usepackage{amsmath}
\usepackage{amssymb}
\usepackage{booktabs}
\usepackage{caption}
\usepackage{url}
\usepackage{graphics,multirow,textcomp,subfigure,multirow,xspace,arydshln,cite}
\usepackage[pagebackref=true,breaklinks=true,letterpaper=true,colorlinks,bookmarks=false,linkcolor=black,citecolor=black,urlcolor=blue]{hyperref}
\usepackage[pdftex]{graphicx}

\DeclareGraphicsExtensions{.eps,.pdf,.jpeg,.png}

\iccpfinalcopy 

\def\iccpPaperID{****} 

\begin{document}

\title{Learned Perceptual Image Enhancement}

\author{Hossein Talebi \qquad Peyman Milanfar \\ Google Research \\ Mountain View, CA \\ {\tt\small \{htalebi, milanfar\}@google.com}
}

\maketitle
\thispagestyle{empty}

\begin{abstract}
   Learning a typical image enhancement pipeline involves minimization of a loss function between enhanced and reference images. While $L_1$ and $L_2$ losses are perhaps the most widely used functions for this purpose, they do not necessarily lead to perceptually compelling results. In this paper, we show that adding a learned no-reference image quality metric to the loss can significantly improve enhancement operators. This metric is implemented using a CNN (convolutional neural network) trained on a large-scale dataset labelled with aesthetic preferences of human raters. This loss allows us to conveniently perform back-propagation in our learning framework to simultaneously optimize for similarity to a given ground truth reference and perceptual quality. This perceptual loss is only used to train parameters of image processing operators, and does not impose any extra complexity at inference time. Our experiments demonstrate that this loss can be effective for tuning a variety of operators such as local tone mapping and dehazing. \end{abstract}

\section{Introduction}

Artificial neural networks have shown promise for image enhancement tasks. Supervised learning has been the common practice among these methods. The main goal in supervised learning is to learn a mapping from input image to ground truth output. Image denoising \cite{chen2017trainable}, deblurring \cite{xu2014deep}, super-resolution \cite{dong2016image}, and tonal adjustments \cite{yan2016automatic} are examples of this generic paradigm. These algorithms rely on training a convolutional neural network (CNN) architecture, with respect to a loss function. Despite recent improvements in image enhancement CNNs, training a deep model with minimal perceptual distortion remains challenging. For the most part, the visual artifacts are due to training with non-perceptual loss functions such as $L_p$ norm. Although $L_1$ and $L_2$ losses are optimization friendly, they tend to result in perceptually inferior solutions.

Perhaps the most widely used perceptual metric for measuring similarity of two images is SSIM (structural similarity) \cite{WangSSIM}. Similar to the human visual system, the SSIM score is based on spatial variations of local image structure. Multi-scale SSIM \cite{wang2003multiscale} is an extension of SSIM which is computed at multiple image scales. This allows for a metric that is less sensitive to image resolution and viewing condition. Recently, Zhao et al. \cite{zhao2017loss} proposed a perceptual loss that consists of multi-scale SSIM and $L_1$ norm, that shows improvement over $L_2$ loss. These loss functions, however, require a reference image, and lack implicit prior information on perceptual image quality. 

Gatys et al. \cite{gatys2015texture, gatys2015neural} proposed a style reconstruction loss for penalizing differences in texture and color. Their loss function is based on a deep pre-trained object detection CNN, and is consequently differentiable. The main idea of this loss is to measure distance between images in the feature space defined by activations of the object detection CNN. Similar to \cite{gatys2015texture, gatys2015neural}, Johnson et al. \cite{johnson2016perceptual} use a CNN feature representation to train feed-forward networks for style transfer and super-resolution tasks. These type of losses have shown significant success for image transformation, and yet, their application in per-pixel optimization remains to be investigated. 

\subsection{Our contributions}
\begin{figure*}[!t]
\vspace{-0 mm}
\begin{center}
\includegraphics*[scale=0.155]{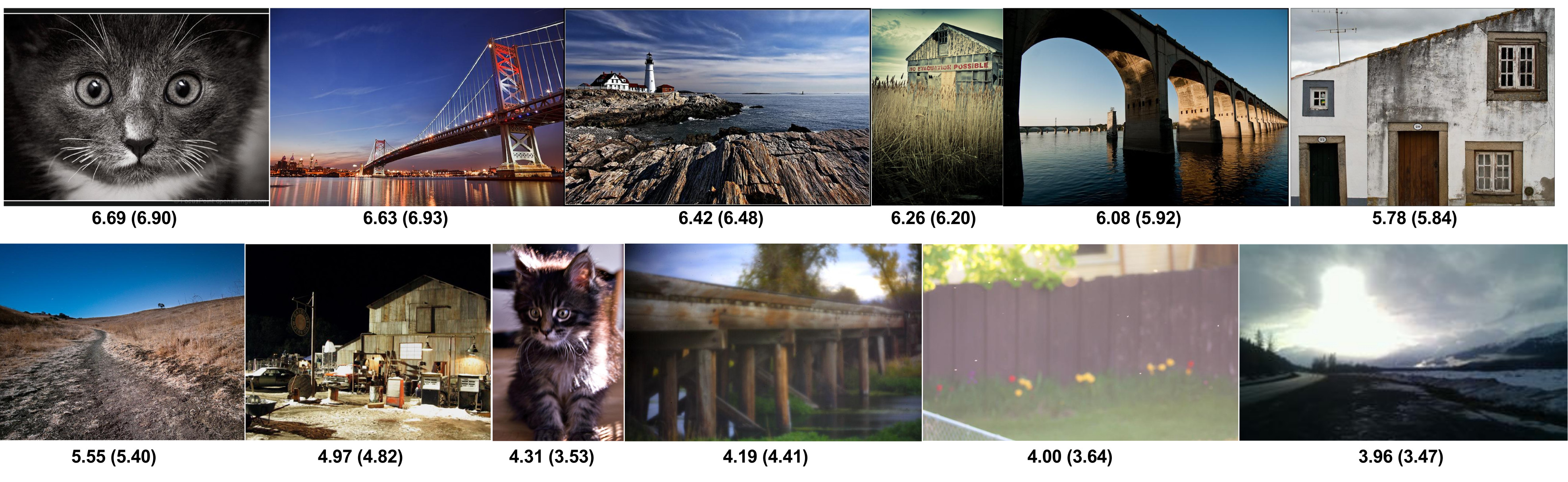}
\end{center}
\vspace{-6 mm}
{\caption{A few examples from the AVA dataset~\cite{murray2012ava}. Our predicted NIMA quality scores are shown below each image (more details in sec.~\ref{sec:nima}). Mean ground truth ratings from AVA are represented in parenthesis. Range of scores in [1,10]. \label{fig:ava_photos}}}
\vspace{-2 mm}
\end{figure*}

This work has two main contributions: 
\begin{enumerate}
\item First, a state-of-the-art no-reference quality predictor is presented that encompasses several aspects of human perceptual preferences. We develop this metric by training a deep neural image assessment (NIMA) model to learn aesthetics and quality of photos from a large scale dataset. Examples of NIMA scores are shown in Fig.~\ref{fig:ava_photos}.
\item Then, NIMA is used as a perceptual loss for image enhancement tasks. We focus on automatic enhancement of lighting, color, tone, contrast and sharpness of images, and show that image enhancement algorithms can effectively benefit from our perceptual loss.
\end{enumerate}

\begin{figure}[!t]
\vspace{-0 mm}
\begin{center}
\includegraphics*[scale=0.135]{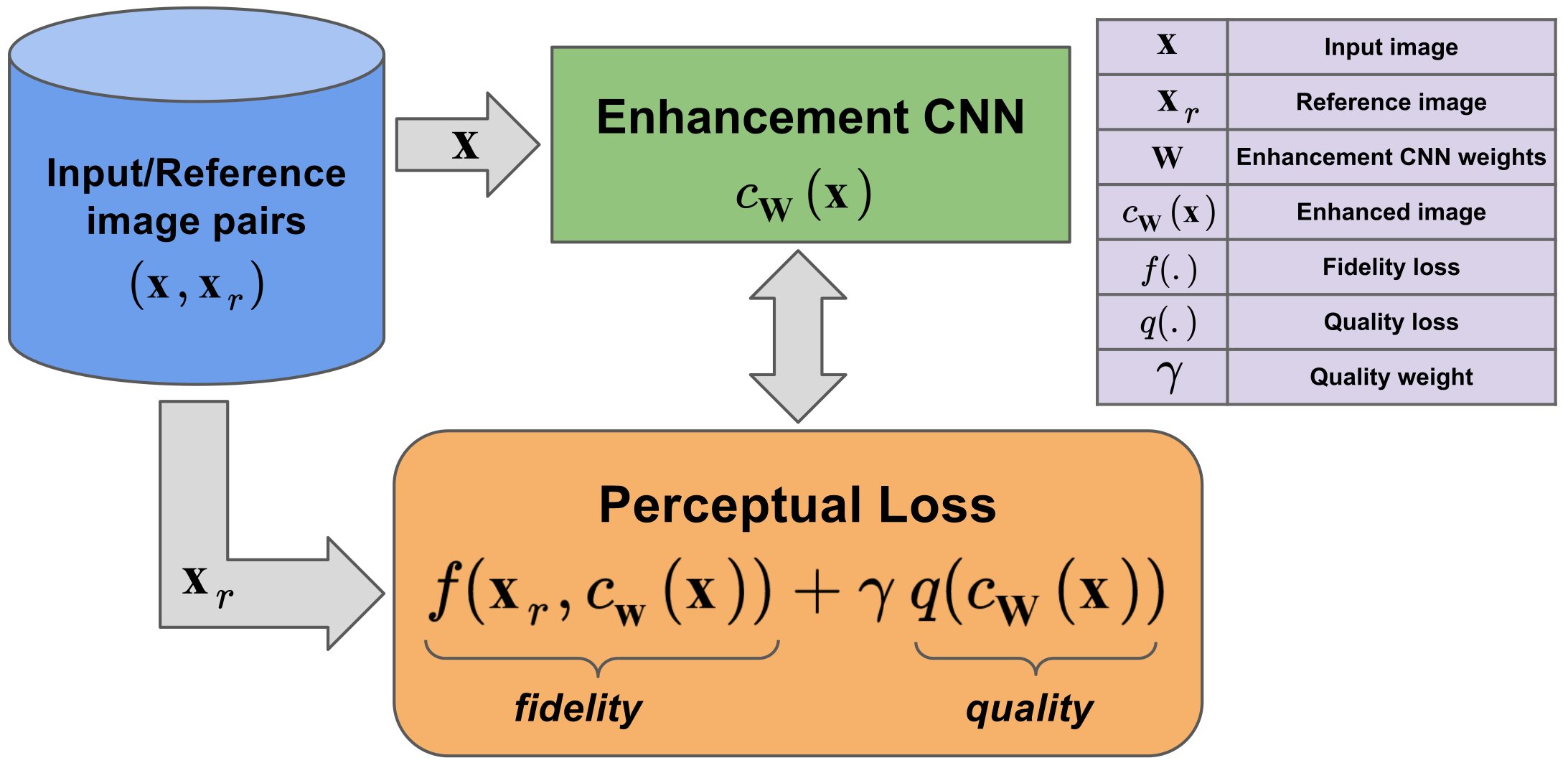}
\end{center}
\vspace{-6 mm}
{\caption{Diagram of our perceptual enhancement training. Training pairs ($\textbf{x}, \textbf{x}_r$) are generated by applying an image processing operator on MIT-Adobe dataset \cite{fivek}. The enhancement CNN is a context aggregation network (CAN) \cite{yu2015multi, chen2017fast}, which generates enhanced image $c_{\textbf{W}}(\textbf{x})$, in which weights $\textbf{W}$ are trained by our perceptual loss. Proposed loss consists of a data fidelity term $f(.)$, and an image quality term $q(.)$. \label{fig:diagram}}}
\vspace{-2 mm}
\end{figure}

\noindent Our proposed framework for training an image enhancement network is shown in Fig.~\ref{fig:diagram}. The proposed loss function consists of a data fidelity term and a perceptual score based on NIMA. Compared to $L_2$ loss, our perceptual loss provides more visually compelling approximation of image processing operators. Our experiments suggest that training an enhancement CNN with the proposed loss results in perceptually superior detail, tone and color in photos. This means that our quality score shows a meaningful improvement when trained with the proposed loss. Next, the large scale dataset used for developing our image quality prior is discussed.

 \subsection{A Large-Scale Database for Aesthetic Visual Analysis (AVA) \cite{murray2012ava}}
 Murray et al.~\cite{murray2012ava} is the benchmark on aesthetic assessment. They introduce the AVA dataset. This dataset consists of 255,000 images, aesthetically rated by amateur photographers. Each image is entered in a single challenge that is associated with a particular contest theme, with nearly 900 different challenges. An average of 200 people scored each photo in response to several photography challenges in all. Raw image ratings range from 1 to 10, with 1 being the lowest aesthetic score. Consequently, a rating histogram with size 10 bins is associated to each image. Mean ratings are concentrated around the overall mean score of 5.5 with a mean standard deviation of 1.4.
 
 A few examples from the AVA dataset are represented in Fig.~\ref{fig:ava_photos}. Visual inspection of the photos in AVA indicates that:
 \begin{itemize}
  \item High quality scores are associated to images with good lighting, tone, contrast and sharpness.
  \item Images with perceptual degradations such as noise and blur are typically rated poorly.
  \item Image semantics play a role in human ratings. Photos that represent visually dull scenery are rated low.
  \item Professional framing and composition can make a photo more appealing to human raters.
\end{itemize}

Modeling these aspects of image quality is not straightforward. Recent learning-based methods \cite{lu2015deep, kao2015visual, jin2016image, mai2016composition} demonstrate a significant performance improvement compared to former works based on hand-crafted features~\cite{murray2012ava}. Lu et al.~\cite{lu2015deep, lu2015rating} propose a double-column CNN that aggregates global and local image views to an overall aesthetic score. Both \cite{murray2012ava} and \cite{lu2015rating} categorize images to only low and high aesthetics based on mean human ratings. An AlexNet inspired  architecture along with a regression loss is used in~\cite{kao2015visual} to predict the mean AVA scores. Similarly, Bin et al.~\cite{jin2016image} retrain a VGG network~\cite{simonyan2014very} to predict the histogram of ratings. Recent work of Mai et al.~\cite{mai2016composition} presents a multi-net approach which extracts image features at multiple scales. Similarly, Ma et al.~\cite{ma2017lamp} use a saliency map to select patches with highest impact on quality score. More recently, Kong et al. in~\cite{kong2016photo} introduced an image ranking method to aesthetically order photos by training with a rank-based loss function. They trained an AlexNet-based CNN to indirectly optimize for rank correlation. Next, our perceptual loss is explained.

\section{Perceptual Loss}

Our proposed loss can be expressed as:
\begin{equation}
\label{eqn:loss}
l(\textbf{W}) = f( \textbf{x}_r, c_{\textbf{W}}(\textbf{x})) + \gamma q(c_{\textbf{W}}(\textbf{x}))
\end{equation}

\noindent where the enhancement network is denoted by $c_\textbf{W}$ with weights $\textbf{W}$, and $\gamma > 0$ controls the strength of the perceptual term. Function $f(.)$ measures a distance between reference image $\textbf{x}_r$ and enhanced image $c_{\textbf{W}}(\textbf{x})$. This function can be seen as a data fidelity term. Typical choices for $f(.)$ are $L_1$, $L_2$ or hybrid $L_1$-$L_2$ losses such as Huber \cite{charbonnier1994two}. The term $q(.)$ is a CNN trained on AVA dataset to predict aesthetic quality of photos. Function $q(.)$ is related to predicted quality score as $q(\textbf{x}) = 10 - \mbox{NIMA}(\textbf{x})$, where $\mbox{NIMA}(\textbf{x})$ is our neural image assessment score for image $\textbf{x}$, and  10 is the highest possible quality score. Next, the NIMA framework is discussed.

\subsection{NIMA: Neural Image Assessment}
\label{sec:nima}

Our image quality predictor is built on image classifier architectures. We explore various classifier architectures such as VGG16~\cite{simonyan2014very}, Inception-v2~\cite{szegedy2016rethinking}, and MobileNet~\cite{howard2017mobilenets}, which are primarily used for object detection. VGG16 consists of 16 layers, 13 convolutional layers with small convolution filters of size $3\times3$, and 3 fully connected layers. By parallel use of convolution and pooling operations, Inception-v2~\cite{szegedy2016rethinking, ioffe2015batch}  provides a more efficient architecture for image classification. MobileNet~\cite{howard2017mobilenets} is another efficient architecture which is primarily designed for mobile vision applications. In MobileNet, convolutional filters are replaced by separable filters which leads to smaller and faster CNN models. Training on the AVA dataset suggests that NIMA architecture based on Inception-v2 produces the best results for predicting the ground truth aesthetic and quality scores.

As shown in Fig.~\ref{fig:nima_architecture}, the last layer of the baseline CNN is replaced with an average pooling layer followed by a fully-connected layer with 10 neurons. Fully-connected layers in all of our baseline CNNs are implemented by convolutional layers. Using fully convolutional layers along with addition of an average pooling (also known as global pooling \cite{he2015spatial}) before the final layer allows us to feed images of arbitrary dimensions to the baseline CNN. This design allows back-propagating from quality score to input pixels. Baseline CNN weights are initialized by training on the ImageNet dataset~\cite{krizhevsky2012imagenet}, and the last fully-connected layer is initialized randomly. All NIMA weights are found by retraining on the AVA dataset. 

\subsubsection{Training NIMA}
\begin{figure}[!t]
\vspace{-0 mm}
\begin{center}
\includegraphics*[scale=0.085]{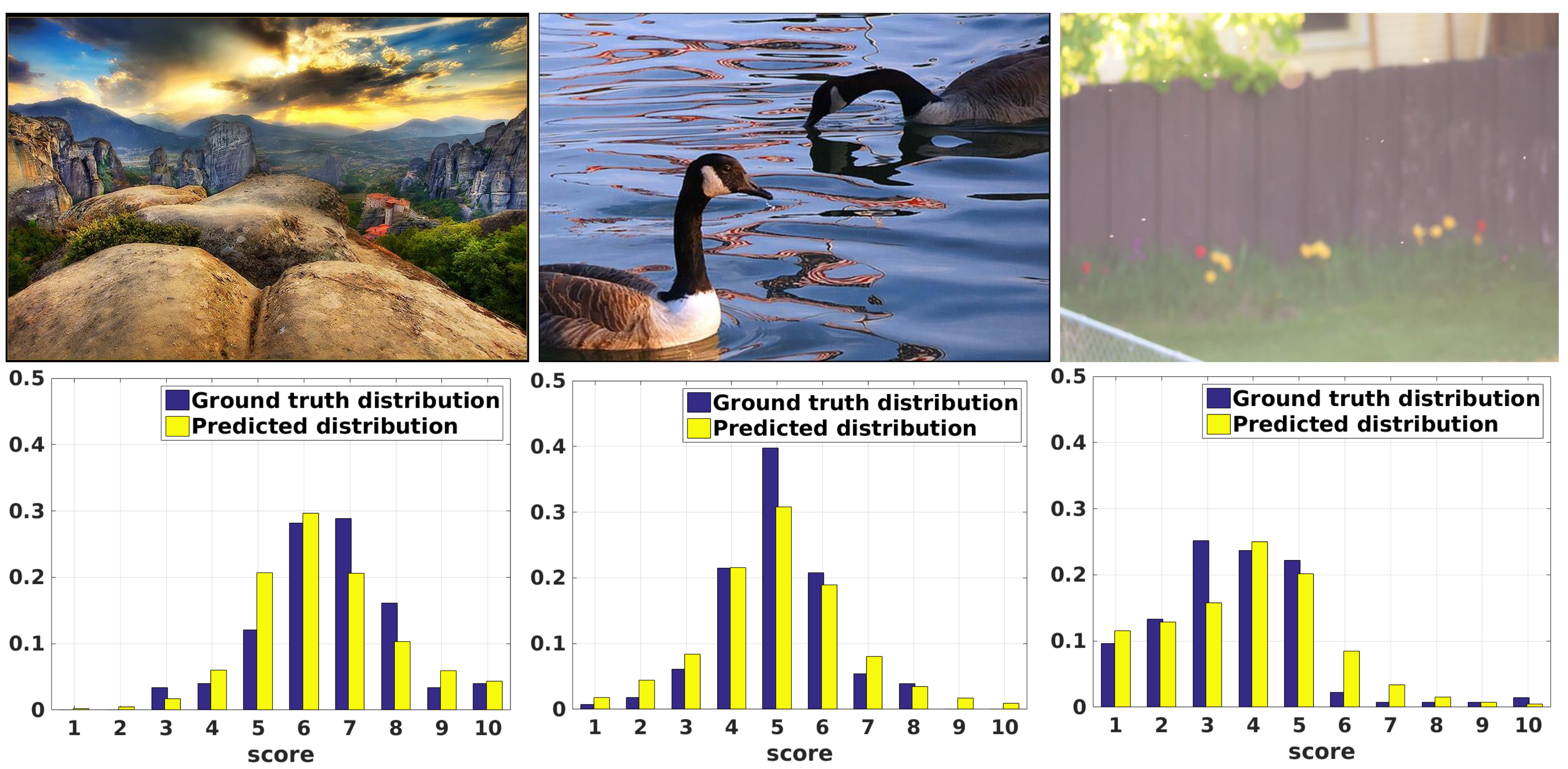}
\end{center}
\vspace{-6 mm}
{\caption{Examples of our predicted quality mass functions for test images from the AVA dataset \cite{murray2012ava}. Corresponding images are shown above each plot. \label{fig:histograms}}}
\vspace{-3 mm}
\end{figure}

In training NIMA, the main objective is to predict the {\em distribution} of quality ratings for a given image. In our notation, ground truth distribution of human ratings of a given image is expressed as an empirical probability mass function $\textbf{p} = [p_{s_1},\ldots,p_{s_N}]$ with $s_1 \leq s_i \leq s_N$, where $s_i$ represents the $i$th score bucket, and $N$ denotes the total number of score buckets. In the AVA dataset, $N=10$, $s_1=1$ and $s_N=10$. Given probability of ratings as $\sum_{i=1}^{N} p_{s_i} = 1$, $p_{s_i}$ represents the probability of a quality score falling in the $i$th bucket. Each AVA example image is assigned a set of ground truth (user) ratings $\textbf{p}$. The objective of our training is to find an accurate estimate of the probability mass function $\textbf{p}$. Some examples of ground truth and predicted probability mass functions are shown in Fig. \ref{fig:histograms}.

\begin{figure}[!t]
\vspace{-0 mm}
\begin{center}
\includegraphics*[scale=0.145]{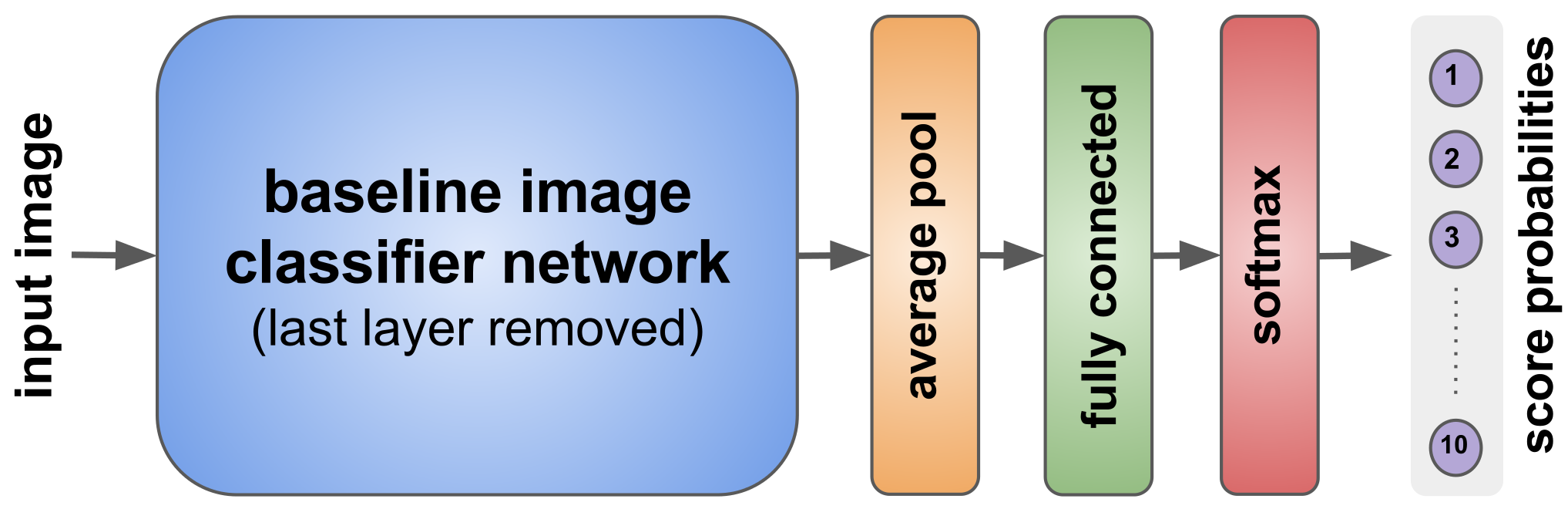}
\end{center}
\vspace{-6 mm}
{\caption{Diagram of our neural image assessment (NIMA) framework. NIMA is built on image classifier architectures. The baseline image classifier network is implemented with fully convolutional layers, and last fully connected layer is replaced with an average pooling followed by a fully connected layer with 10 output neurons to predict quality scores of the AVA dataset \cite{murray2012ava}. \label{fig:nima_architecture}}}
\vspace{-3 mm}
\end{figure}

Given the inter-class relationships between score buckets, we select an EMD-based loss \cite{hou2016squared}  to train the NIMA model. In contrast to cross-entropy loss, EMD loss penalizes mis-classifications according to class distances. Quality rating classes are inherently ordered as $s_1 < \dots  < s_N$, and $l-$norm distance between classes can be expressed as $\|s_i - s_j\|_l$, where $1 \leq i, j \leq N$. EMD is defined as the minimum cost to move the mass of one distribution (ground truth probability mass function $\textbf{p}$) to another (estimated probability mass function $\widehat{\textbf{p}}$). With $N$ ordered classes of distance $\|s_i - s_j\|_l$, the normalized Earth Mover's Distance can be defined as~\cite{levina2001earth}:
\begin{equation}
\label{eqn:emd}
\mbox{EMD}(\textbf{p}, \widehat{\textbf{p}}) = \left( \frac{1}{N} \sum_{k=1}^{N} |\mbox{CDF}_{\textbf{p}}(k) - \mbox{CDF}_{\widehat{\textbf{p}}}(k)|^l \right)^{1/l}
\end{equation}
where the cumulative distribution function $\mbox{CDF}_{\textbf{p}}(k)=\sum_{i=1}^{k} p_{s_i}$. This definition of normalized EMD requires both distributions to have equal mass: $\sum_{i=1}^{N} p_{s_i} = \sum_{i=1}^{N} \widehat{p}_{s_i}$. As shown in Fig.~\ref{fig:nima_architecture}, our predicted score probabilities are passed through a soft-max function to guarantee that $\sum_{i=1}^{N} \widehat{p}_{s_i} = 1$. Similar to~\cite{hou2016squared}, in our training framework, $l$ is set as 2 to allow easier optimization when working with gradient descent.

It is worth mentioning that the mean score obtained from estimated mass function $\widehat{\textbf{p}}$ is used as quality score in our perceptual loss function (\ref{eqn:loss}). More explicitly, $\mbox{NIMA}(\textbf{x})=\sum_{i=1}^{10} s_i \widehat{p}_{s_i}$, where $\widehat{\textbf{p}}$ is associated with image $\textbf{x}$. (examples of $\widehat{\textbf{p}}$ are shown in Fig. \ref{fig:histograms}).
 
\section{Enhancement CNN}
\begin{figure}[!t]
\vspace{-0 mm}
\begin{center}
\includegraphics*[scale=0.12]{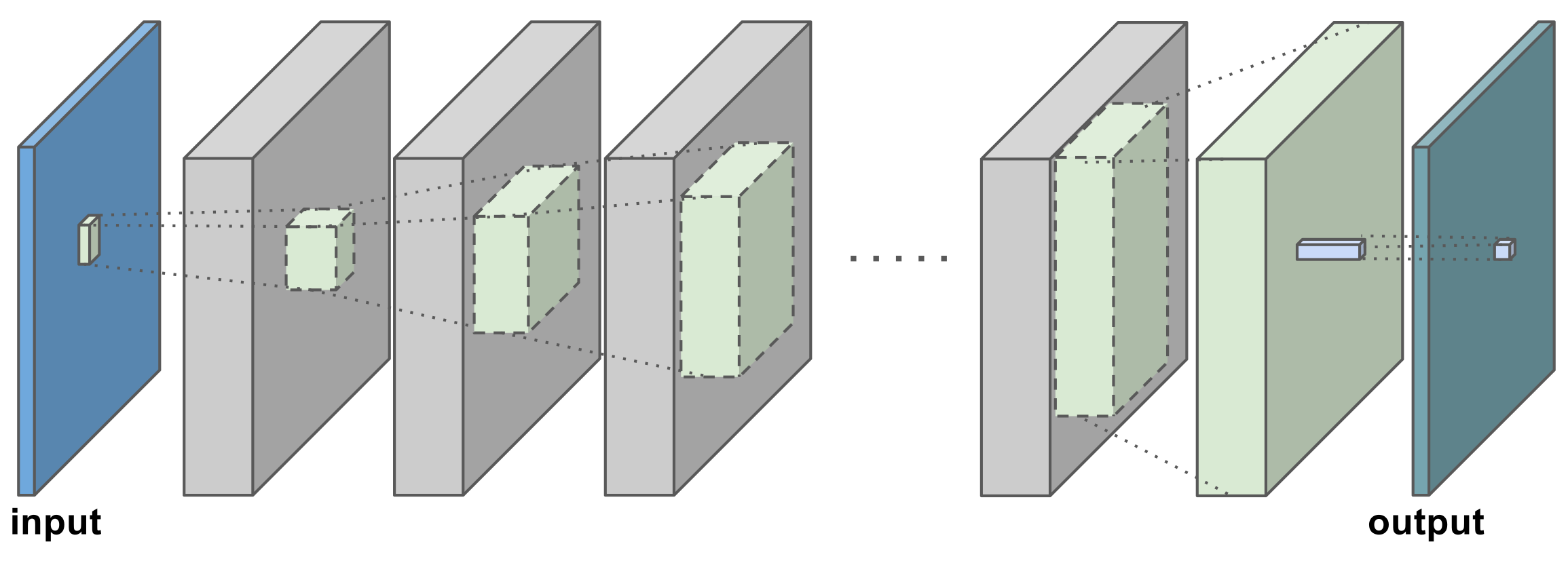}
\end{center}
\vspace{-6 mm}
{\caption{Effective receptive fields of context aggregation network (CAN) architecture \cite{yu2015multi} used as our enhancement CNN. CAN uses dilated convolutions with increasing dilation rates in depth, and consequently, each output pixel becomes an aggregation of several input pixels. \label{fig:can_architecture}}}
\vspace{-3 mm}
\end{figure}

Our enhancement architecture is the multi-scale context aggregation network (CAN), developed primarily for semantic segmentation \cite{yu2015multi}. This architecture is shown in Fig. \ref{fig:can_architecture}. CAN uses dilated convolutions to aggregate global contextual information. By an exponential expansion of filter receptive fields in the dilated convolution, the CAN architecture allows feeding images with arbitrary size without need to rescale. This CNN was recently used by Chen et al. \cite{chen2017fast} to approximate a few image processing operations such as edge-aware filtering, local tone and detail manipulation, dehazing, and photographic style transfer. Visual inspection of results in \cite{chen2017fast} indicates that the CAN architecture is more suitable for approximating global operations such as tone and contrast enhancement. This is not surprising because dilated CNNs with progressive receptive fields tend to be global operators. In other words, by dilating filters in CNN's depth, a large number of pixels contribute to compute an output pixel. Since we are aiming to optimize for best attainable global enhancements, CAN suits our application.

Similar to \cite{chen2017fast}, we use the CAN architecture with 10 convolutional layers with 32 feature maps, and kernels of size $3\times3$ for intermediate layers and $1\times1$ for the last layer. The dilation rate is increased exponentially as $2^{k}$ for layer $k$, where $0 \leq k < d-1$, and $d=10$ total number of convolutional layers. Layer $d-1$ is not dilated. We use leaky rectified linear unit \cite{maas2013rectifier} as nonlinearities in all convolutional layers, except the last layer which has no nonlinearity. Unlike \cite{chen2017fast}, our intermediate feature maps are symmetrically padded. This is an essential modification to avoid artifacts created by zero padded layers with high dilation rates. Next, we represent our experimental results with CAN, trained with the proposed perceptual loss to address the aforementioned issues.

\section{Experimental Results}

In this section, first we train NIMA models with various architectures on the AVA dataset. Our results indicate that the proposed quality predictor models are comparable to state-of-the-art methods, but have much lower complexity. Then, results for training the enhancement CNN with NIMA as the training loss are discussed. The CNNs presented in this work are implemented using TensorFlow~\cite{abadi2016tensorflow}.

\subsection{NIMA Results}
The AVA dataset is divided to 80\% training and 20\% test sets. All images are scaled to $480\times 640$ resolution, which is roughly the average resolution of AVA images (largest image dimension is scaled to $640$). The baseline CNN is initialized by training on ImageNet~\cite{krizhevsky2012imagenet}, and the last fully-connected layer is randomly initialized. Using a momentum optimizer, the learning rate of the baseline CNN layers and the last fully-connected layers are set as $3\times10^{-7}$ and $3\times 10^{-6}$, respectively. Also, after every 10 epochs of training with batch size 64, an exponential decay is applied to the learning rates.

Evaluation of NIMA models on the test set vs. other methods are presented in Table~\ref{tab:ava_comp}. Most existing methods in Table~\ref{tab:ava_comp} are designed to perform binary classification on AVA scores. Therefore, only accuracy values of two-class quality categorization (low/high quality) are reported. Results from~\cite{ma2017lamp}, and NIMA(Inception-v2) show the highest accuracy. In terms of rank correlation, NIMA(VGG16) and NIMA(Inception-v2) outperform~\cite{kong2016photo}. It is worth noting that \cite{ma2017lamp} applies multiple VGG16 nets on image patches to predict a single quality score, whereas NIMA(Inception-v2) requires only one pass of the Inception-v2 network. Predictions of NIMA(Inception-v2) are represented in Fig. \ref{fig:ava_photos}. As can be seen, ground truth ranking of these photos is closely predicted by the NIMA score. Also, the computational complexities of the different NIMA models are presented in Table~\ref{tab:time_comp}. Given the performance comparison and complexity trade-offs, NIMA(Inception-v2) model is used as a loss to train the enhancement task. 

\begin{table}[!t]
\begin{center}
\caption{Performance of proposed neural image assessment (NIMA) with various architectures in predicting AVA quality ratings~\cite{murray2012ava} compared to the state-of-the-art. Reported accuracy values are based on classification of photos to two classes (column 2). LCC (linear correlation coefficient) and SRCC (Spearman's rank correlation coefficient) are computed between predicted and ground truth mean scores (column 3 and 4). EMD measures closeness of the predicted and ground truth rating distributions with $r=1$ in Eq. \ref{eqn:emd}.}
\begin{tabular}{@{} *5l @{}}    \toprule
\emph{Model} & \emph{Accuracy} & \emph{LCC} & \emph{SRCC} & \emph{EMD} \\\midrule
Murray et al.~\cite{murray2012ava}   & 66.70\% &  --  &  -- &  --   \\ 
Kao et al.~\cite{kao2015visual}    & 71.42\% &  --  &  --  &  --  \\ 
Lu et al. \cite{lu2015rating}    & 75.42\%  & -- &  --  &  -- \\ 
Kao et al. \cite{kao2016visual}    & 76.58\% & -- &  --  &  --  \\ 
Wang et al. \cite{wang2016brain}    & 76.80\% & -- &  --  &  -- \\ 
Mai et al. \cite{mai2016composition}    & 77.10\%  & -- &  -- &  --    \\ 
Kong et al. \cite{kong2016photo}  &  77.33\% & --  & 0.558 &  --   \\
Ma et al. \cite{ma2017lamp} & 81.70\% & -- &  -- &  --   \\ \hdashline
NIMA(MobileNet)  & 80.71\%  & 0.565 & 0.534 &  0.070 \\
NIMA(VGG16)  & 80.96\%  & 0.631 & 0.605 &   0.051 \\
NIMA(Inception-v2)  & \textbf{81.88}\%  & \textbf{0.660} & \textbf{0.636} &  \textbf{0.048} \\\bottomrule
 \hline
\end{tabular}
\label{tab:ava_comp}
\end{center}
\vspace{-6 mm}
\end{table}

\subsection{Enhancement Results}

The enhancement CNN is trained and evaluated on the MIT-Adobe FiveK dataset \cite{fivek}. We split the FiveK dataset to training and testing sets with 2500 images. At training, images are rescaled to $480\times 640$. However, testing is performed on the original size of images. Also, pixel values are mapped to [0,1]. We obtain two sets of reference images by applying the tone enhancement operator of \cite{Sylvain2011}, and nonlocal image dehazing of \cite{NonLocalImageDehazing} on the FiveK dataset.

We use an $L_2$ loss as our data fidelity function in (\ref{eqn:loss}); however, any other differentiable full-reference loss can be used.  The perceptual $\gamma$ is set as 0.0001. It is worth noting that since our perceptual quality measure is a no-reference metric, large values of $\gamma$ may cause unexpected distortions in the output. Conversely, small values of $\gamma$ barely lead to visible improvement of results from the baseline $L_2$ loss. We trained several enhancement models to find the optimal value of $\gamma$. For performance comparison, we trained the same model with $\gamma=0$, which is equivalent to results from \cite{chen2017fast}.

We select the Adam optimizer \cite{kingma2014adam} with learning rate set as $0.0001$, and batch size as 1. CAN is trained for $5\times10^{6}$ steps of stochastic gradient descent. Weights from NIMA network are kept fixed during training.

\begin{figure*}[!t]
\vspace{-0 mm}
\begin{center}
\includegraphics*[scale=0.0733]{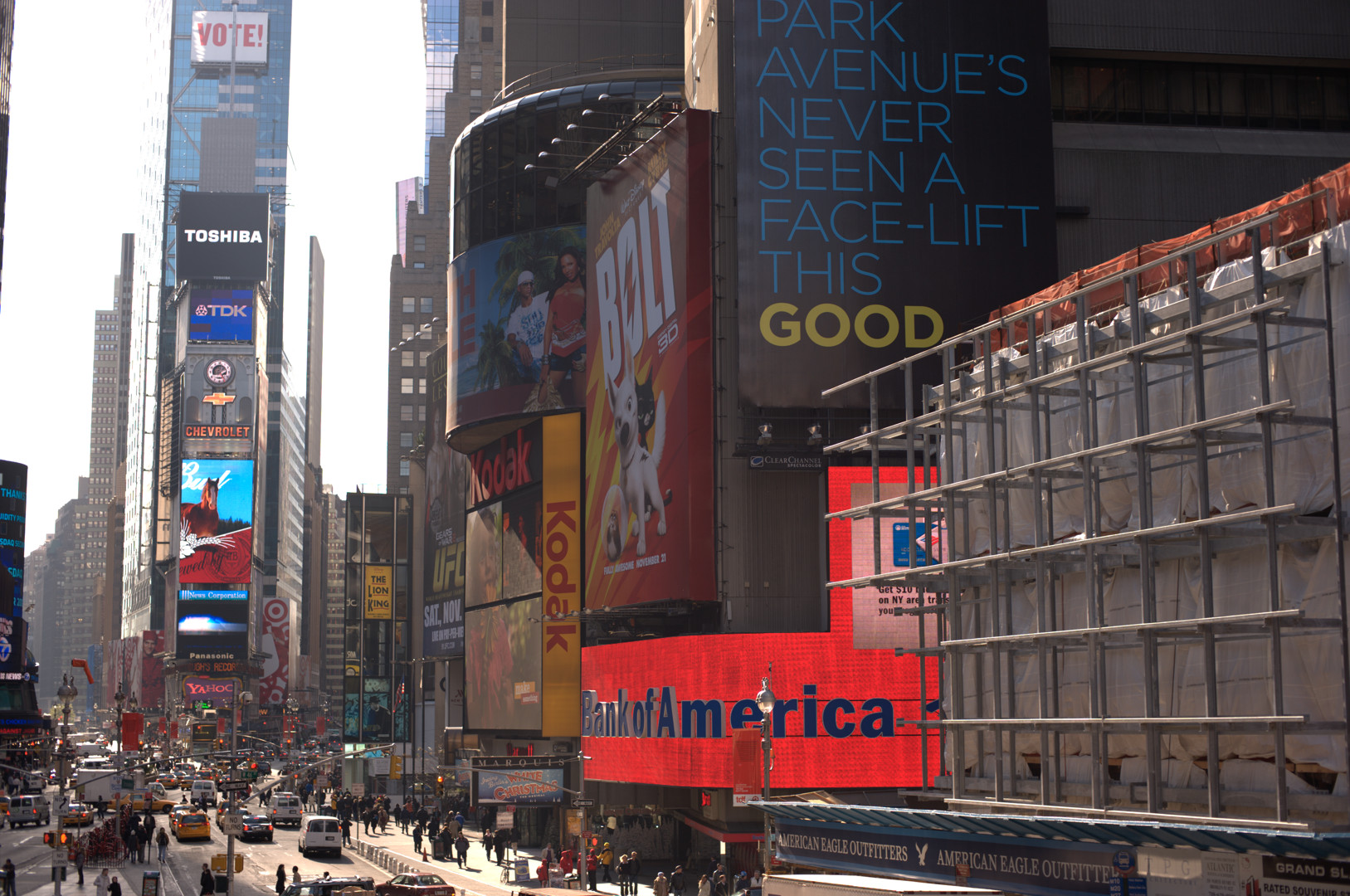}
\includegraphics*[scale=0.0733]{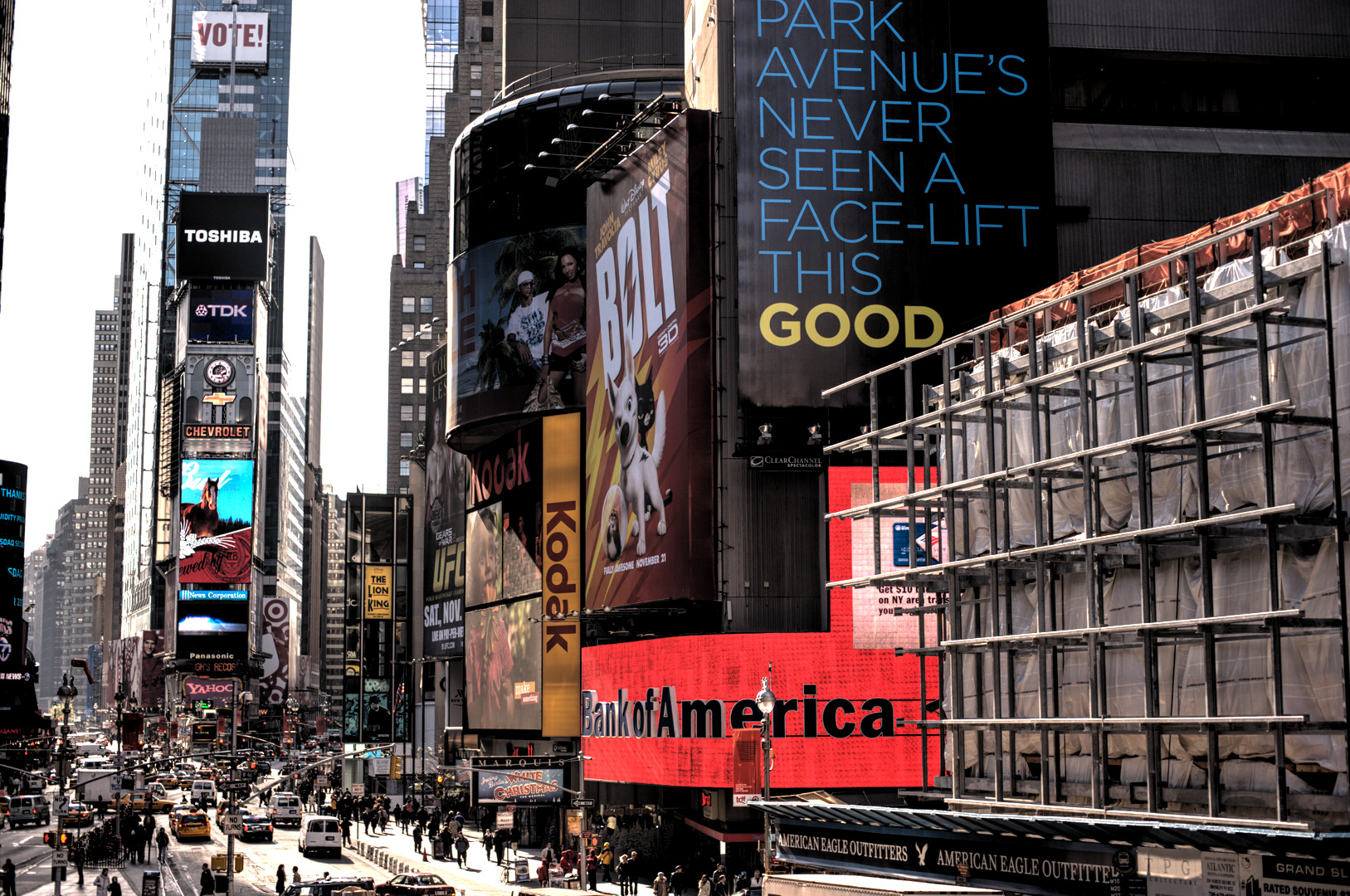}
\includegraphics*[scale=0.0733]{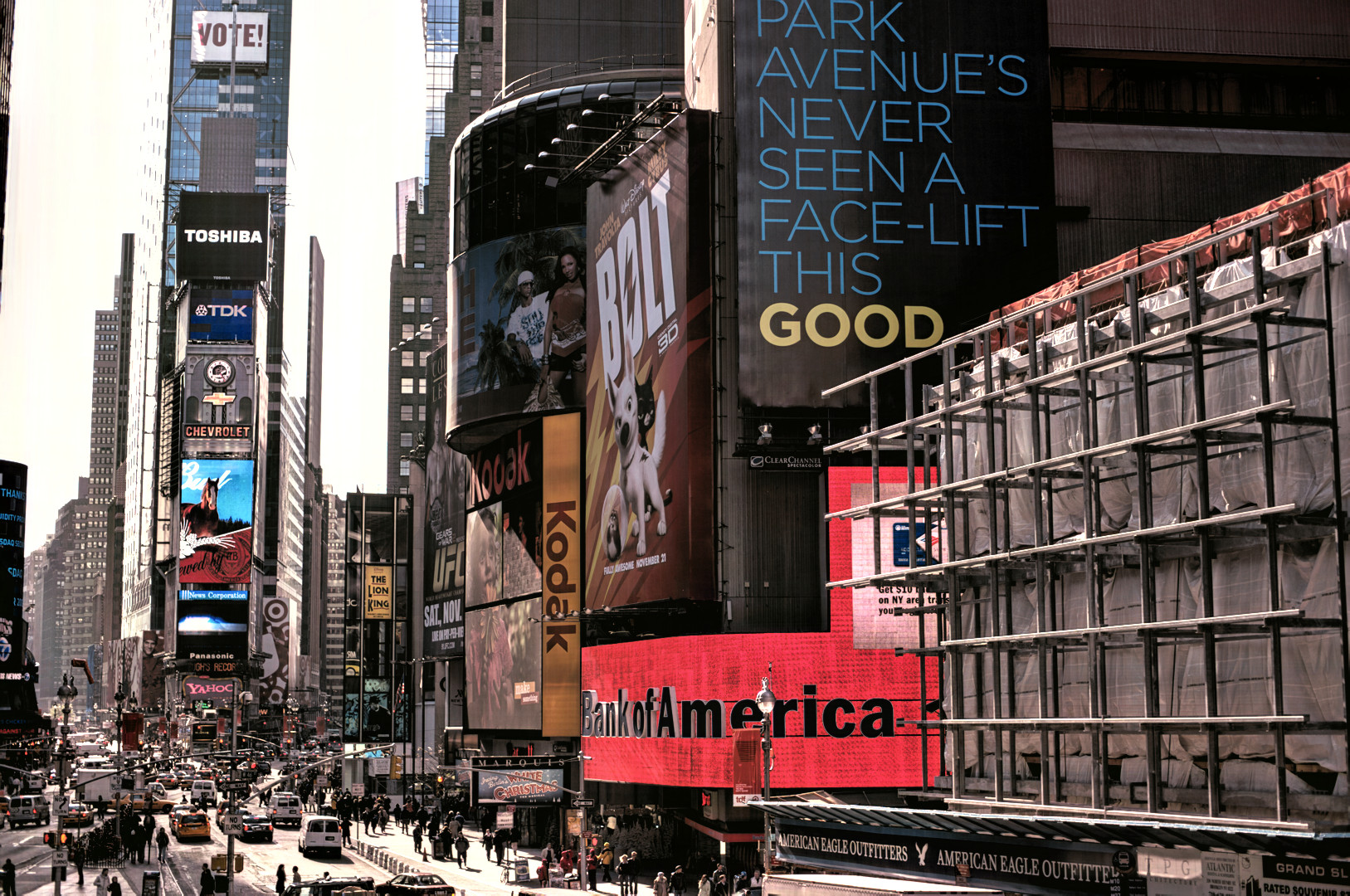}
\includegraphics*[scale=0.0733]{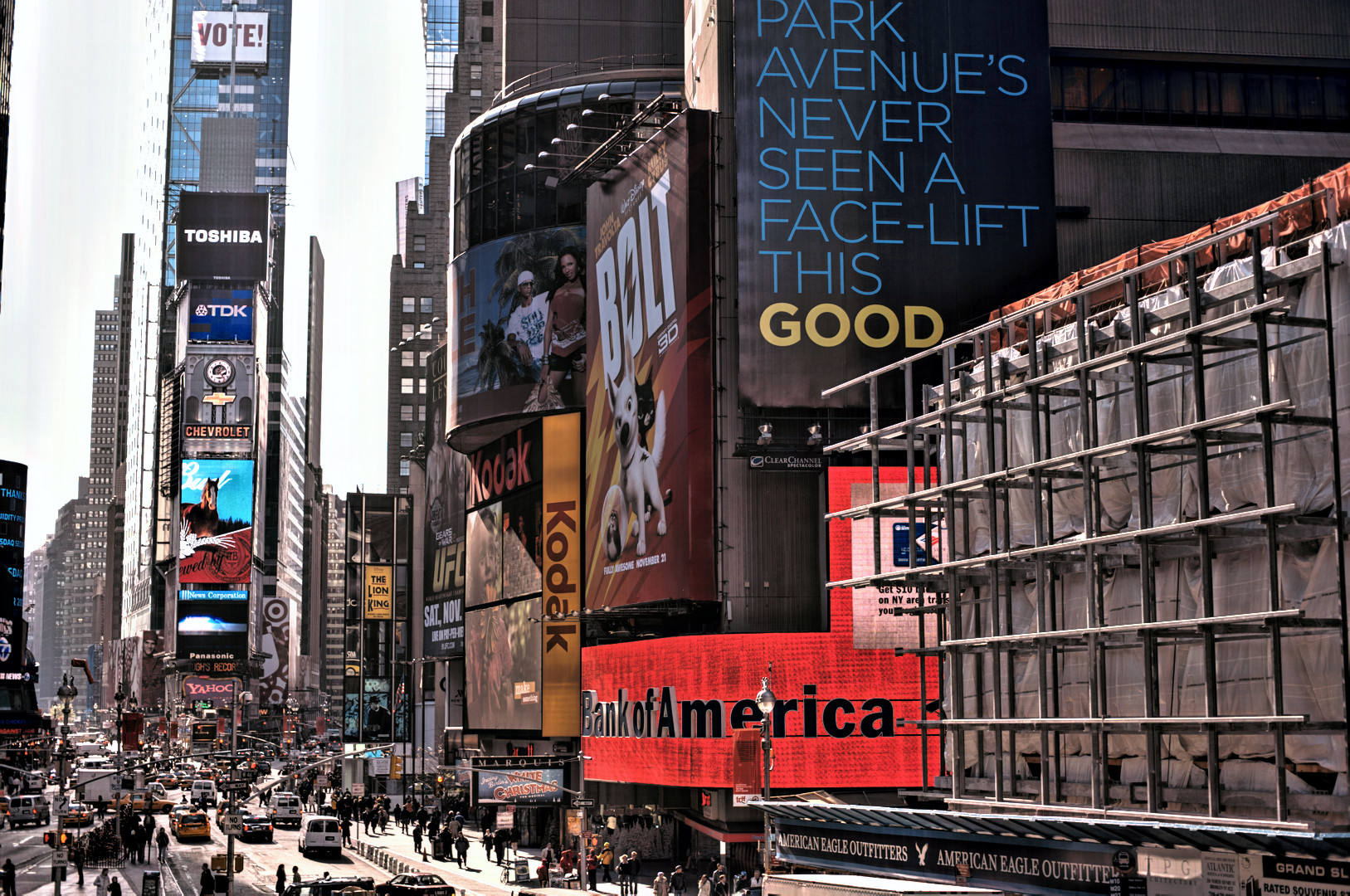}
\includegraphics*[viewport=565 490 640 555, scale=1.59]{./figures/000016.jpg}
\includegraphics*[viewport=565 490 640 555, scale=1.59]{./figures/000016_detail_manip_gt.jpg}
\includegraphics*[viewport=565 490 640 555, scale=1.59]{./figures/000016_detail_manip_l2.jpg}
\includegraphics*[viewport=565 490 640 555, scale=1.59]{./figures/000016_detail_manip_l2_nima.jpg}
\includegraphics*[viewport=550 790 650 940, scale=1.19]{./figures/000016.jpg}
\includegraphics*[viewport=550 790 650 940, scale=1.19]{./figures/000016_detail_manip_gt.jpg}
\includegraphics*[viewport=550 790 650 940, scale=1.19]{./figures/000016_detail_manip_l2.jpg}
\includegraphics*[viewport=550 790 650 940, scale=1.19]{./figures/000016_detail_manip_l2_nima.jpg}
\includegraphics*[scale=0.0736]{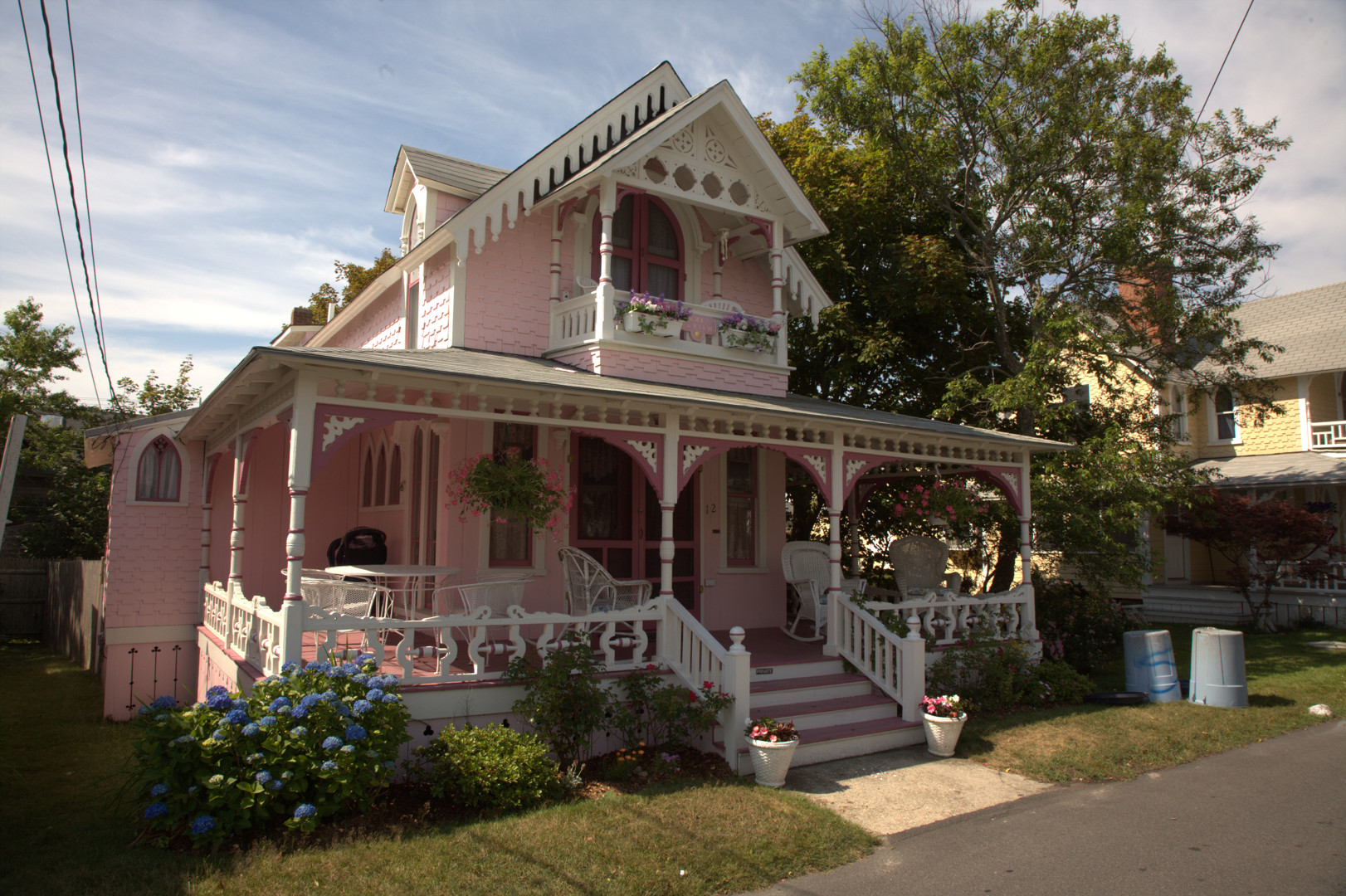}
\includegraphics*[scale=0.0736]{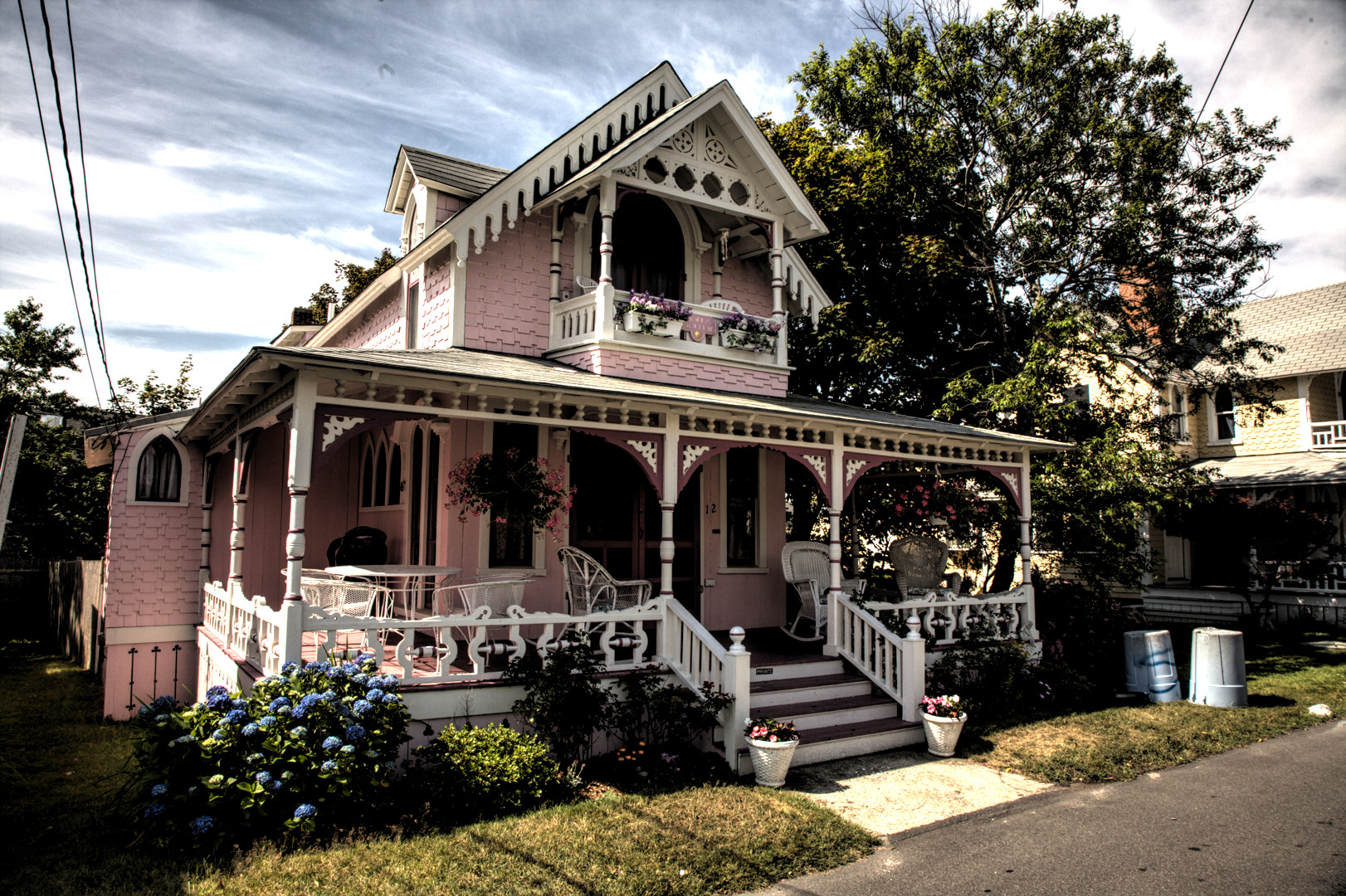}
\includegraphics*[scale=0.0736]{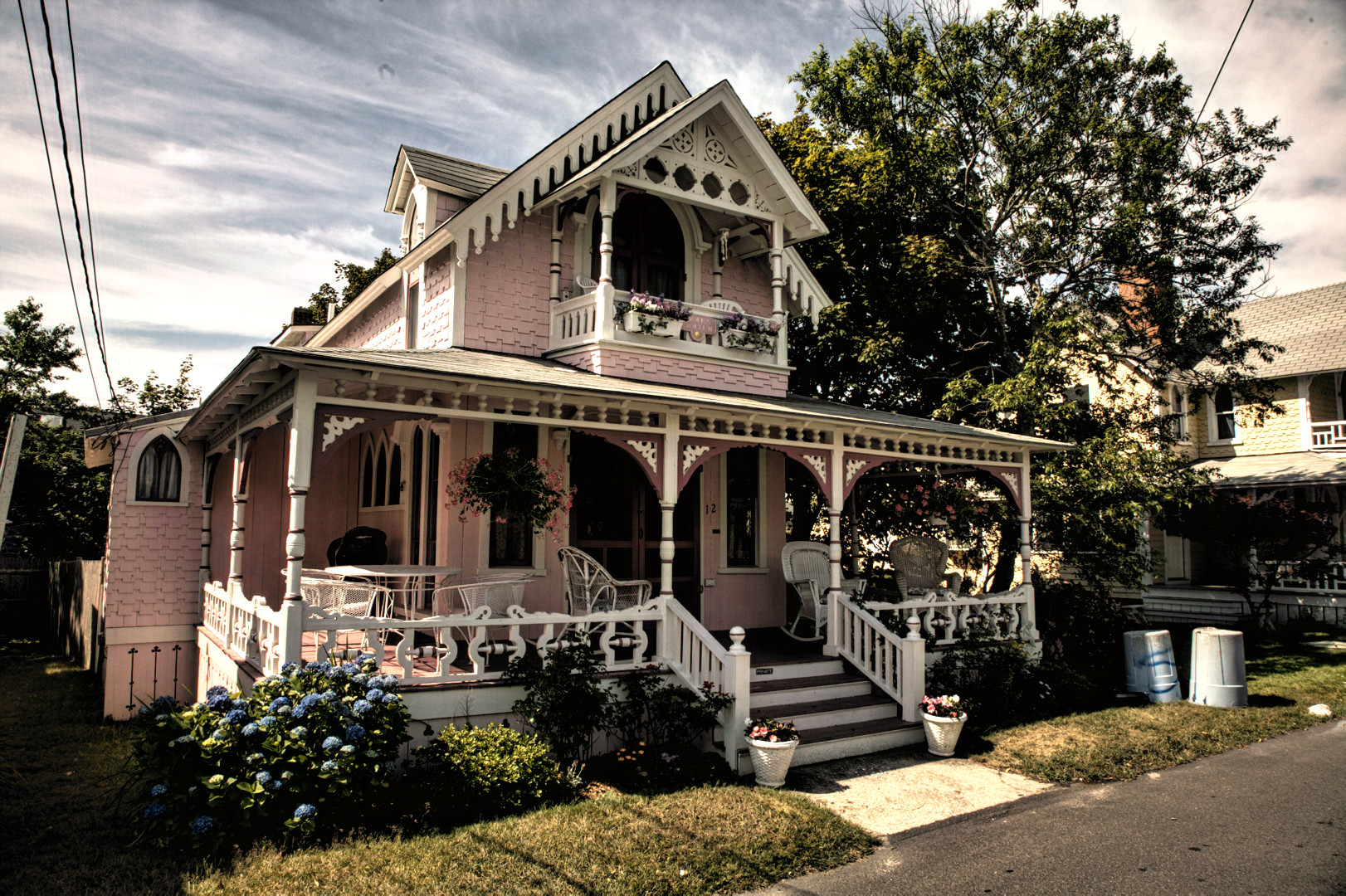}
\includegraphics*[scale=0.0736]{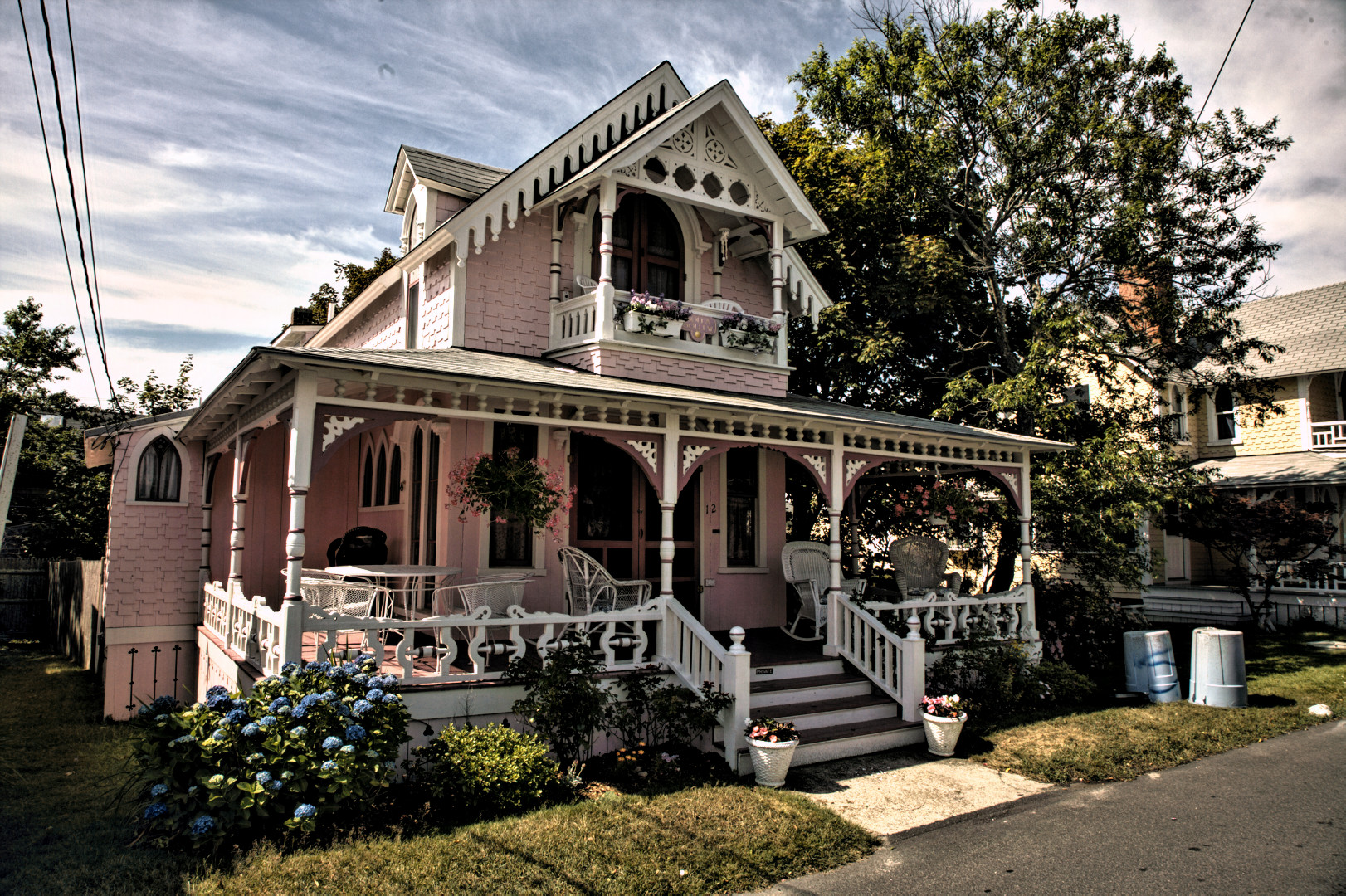}
\\[-1.7mm]
\subfigure[\scriptsize Input]{
\includegraphics*[viewport=700 680 880 830, scale=0.661]{./figures/000178.jpg}} \hspace{-1.5mm}
\subfigure[\scriptsize Reference \cite{Sylvain2011}]{
\includegraphics*[viewport=700 680 880 830, scale=0.661]{./figures/000178_detail_manip_gt.jpg}} \hspace{-1.5mm}
\subfigure[\scriptsize CAN($L_2$) \cite{chen2017fast}]{
\includegraphics*[viewport=700 680 880 830, scale=0.661]{./figures/000178_detail_manip_l2.jpg}} \hspace{-1.5mm}
\subfigure[\scriptsize CAN($L_2$+NIMA)]{
\includegraphics*[viewport=700 680 880 830, scale=0.661]{./figures/000178_detail_manip_l2_nima.jpg}} \hspace{-1.5mm}
\end{center}
\vspace{-8 mm}
{\caption{Comparison of local detail enhancement operators in dark regions. In comparison to local tone mappers in \cite{Sylvain2011} and \cite{chen2017fast}, image details in dark and bright areas are better preserved in our results. \label{fig:dark}}}
\vspace{-4 mm}
\end{figure*}

\begin{figure*}[!t]
\vspace{-0 mm}
\begin{center}
\includegraphics*[scale=0.11]{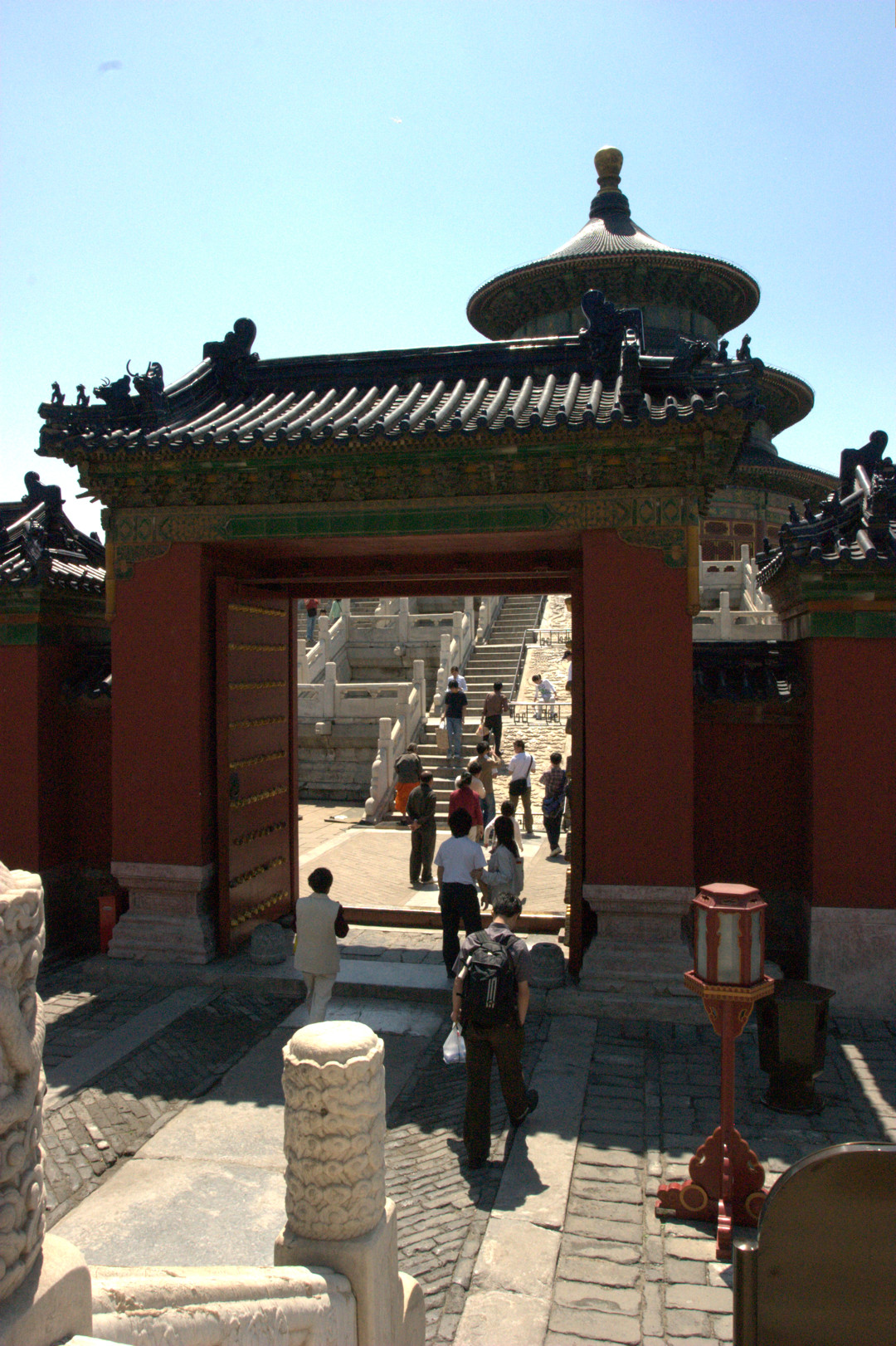}
\includegraphics*[scale=0.11]{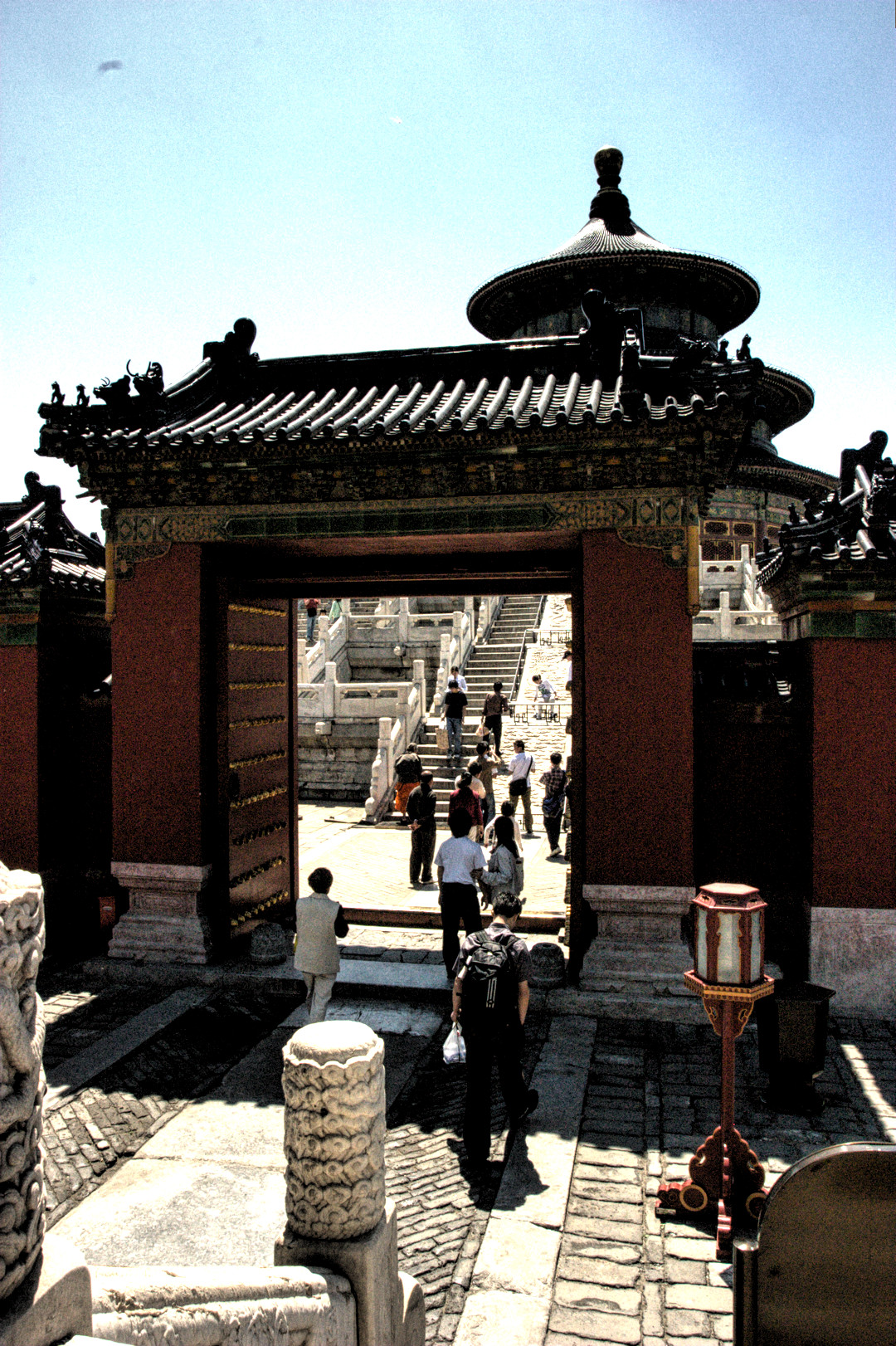}
\includegraphics*[scale=0.11]{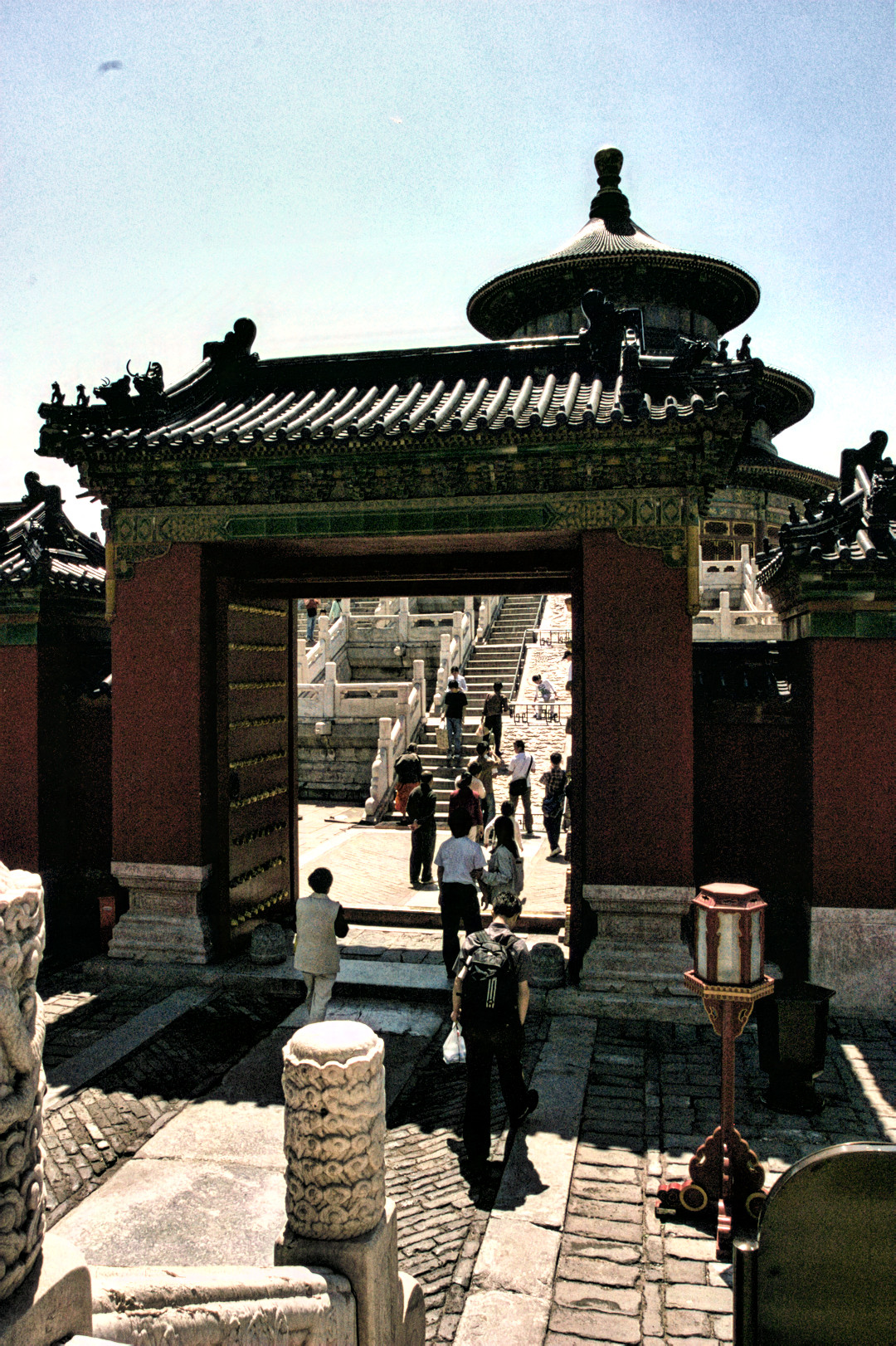}
\includegraphics*[scale=0.11]{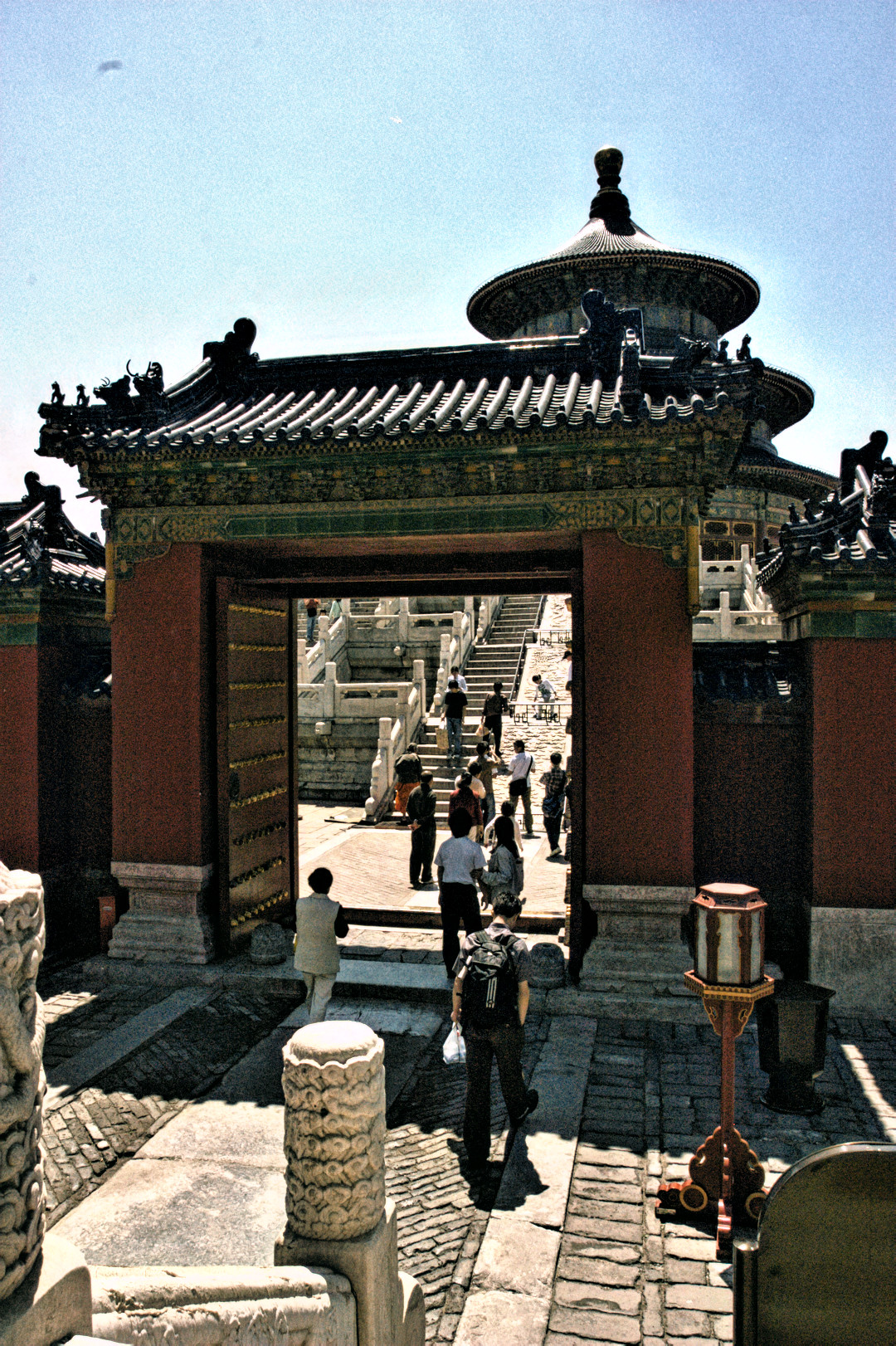}
\includegraphics*[viewport=400 510 570 630, scale=0.7]{./figures/000034.jpg}
\includegraphics*[viewport=400 510 570 630, scale=0.7]{./figures/000034_detail_manip_gt.jpg}
\includegraphics*[viewport=400 510 570 630, scale=0.7]{./figures/000034_detail_manip_l2.jpg}
\includegraphics*[viewport=400 510 570 630, scale=0.7]{./figures/000034_detail_manip_l2_nima.jpg}
\includegraphics*[scale=0.0733]{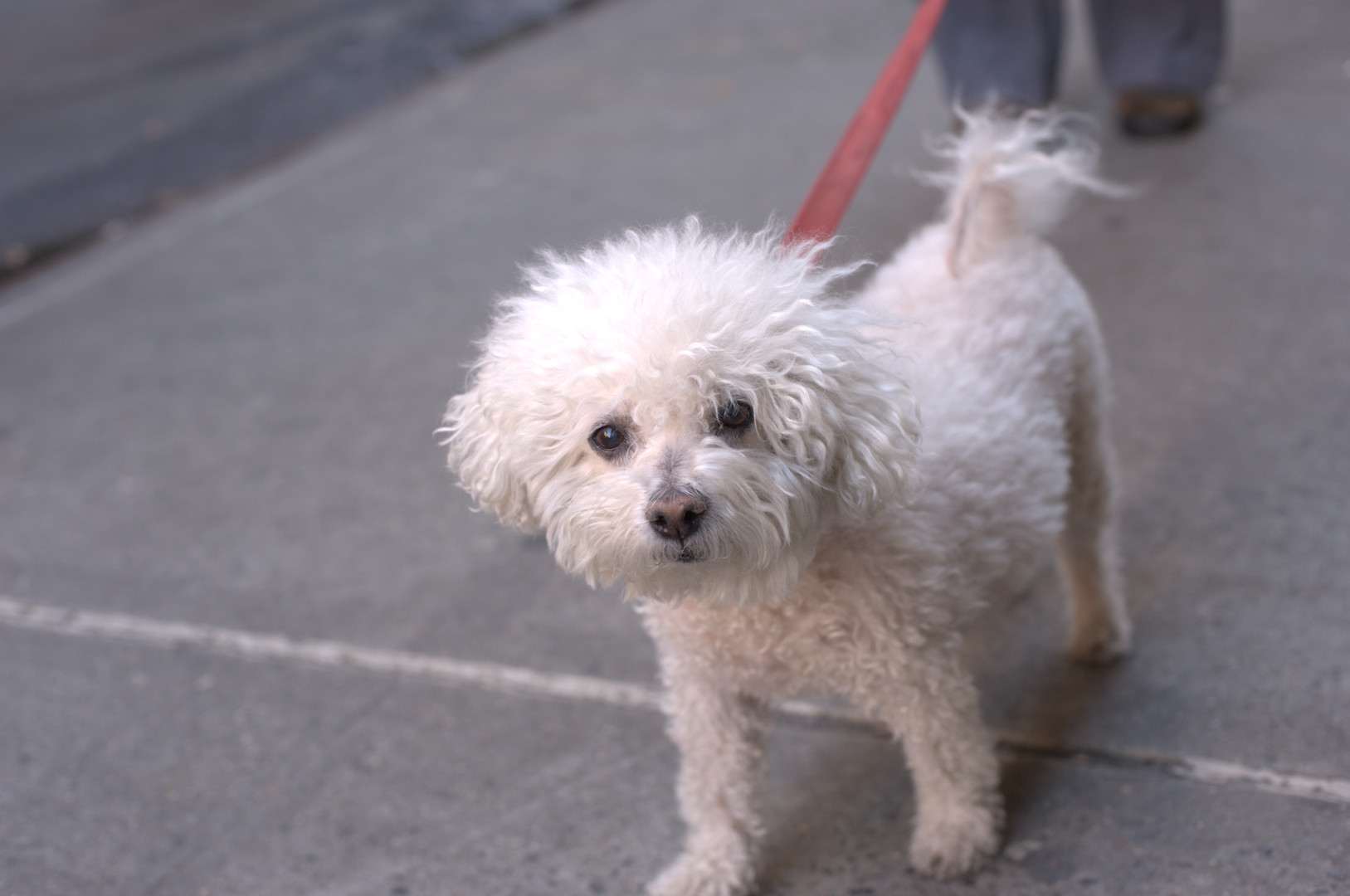}
\includegraphics*[scale=0.0733]{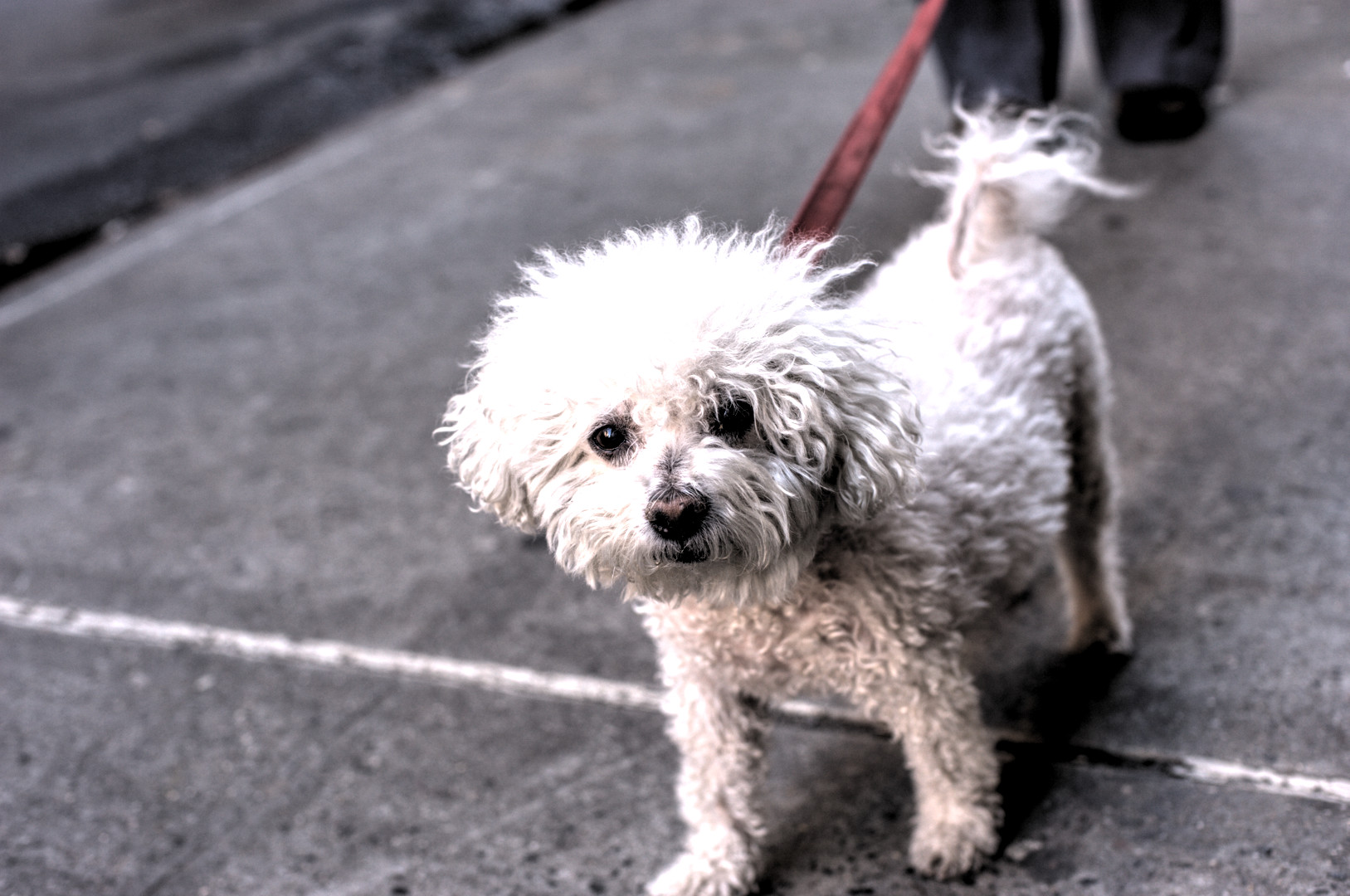}
\includegraphics*[scale=0.0733]{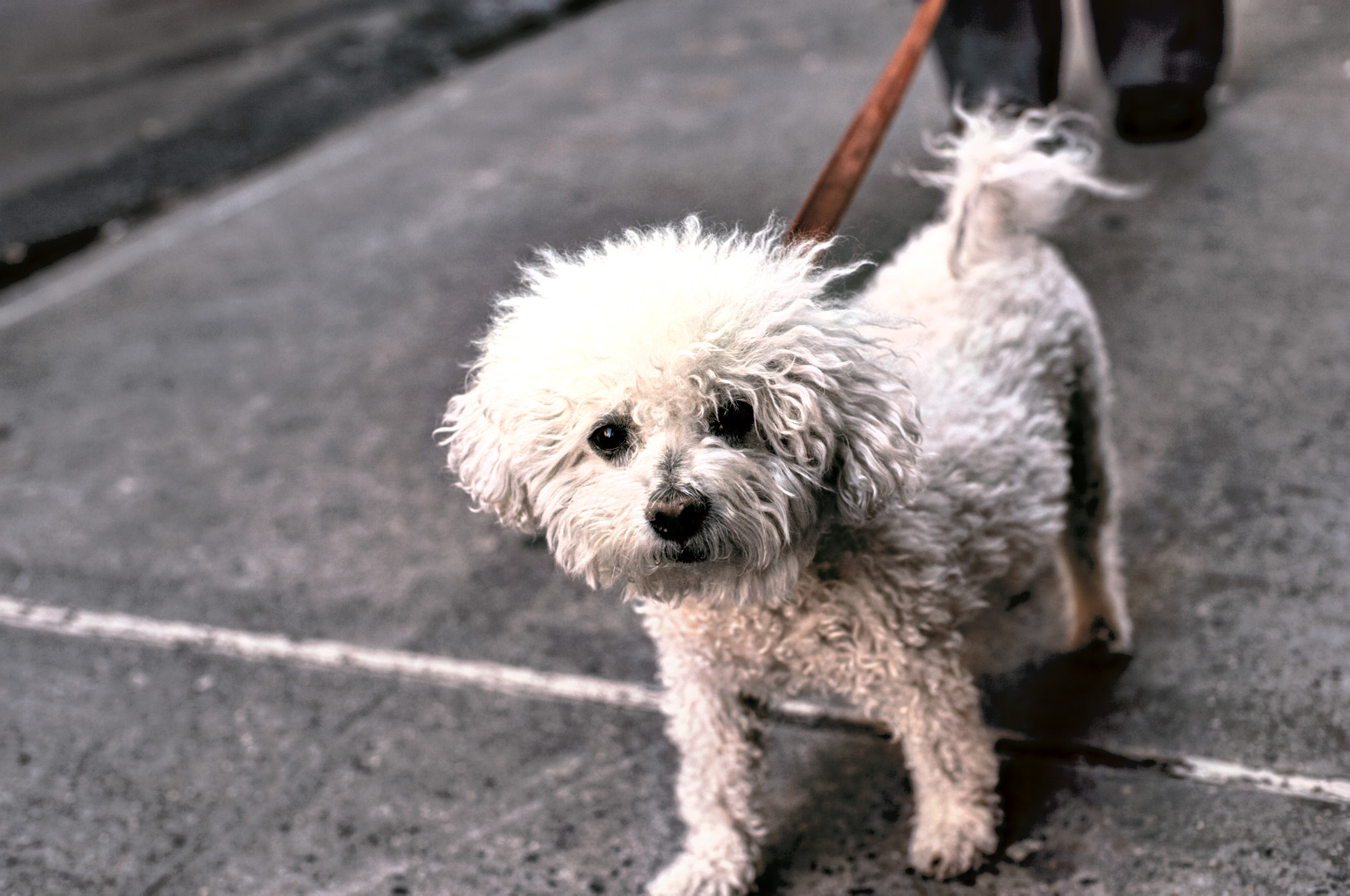}
\includegraphics*[scale=0.0733]{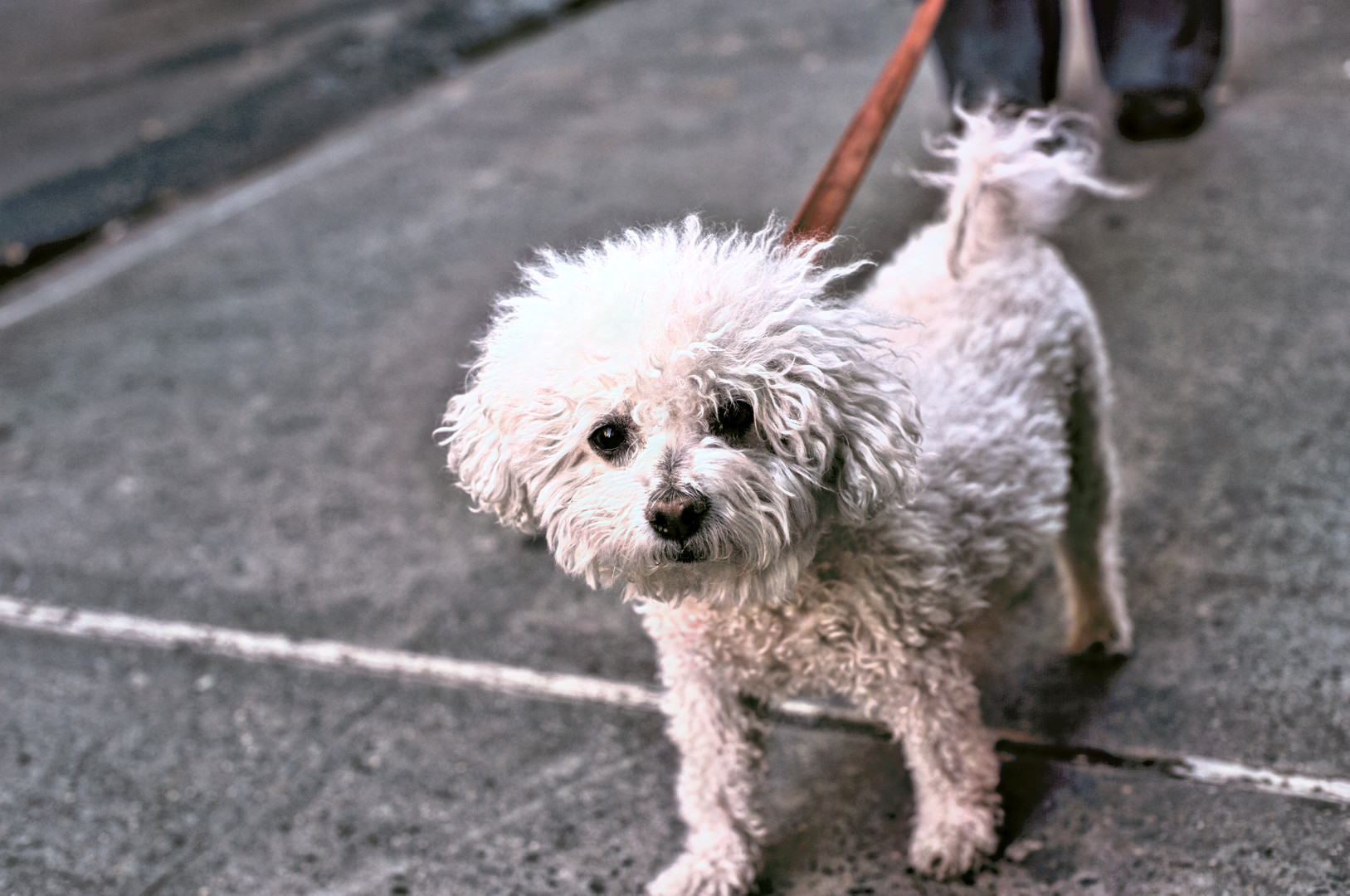}
\\[-1.7mm]
\subfigure[\scriptsize Input]{
\includegraphics*[viewport=800 680 980 830, scale=0.66]{./figures/000102.jpg}} \hspace{-1.5mm}
\subfigure[\scriptsize Reference \cite{Sylvain2011}]{
\includegraphics*[viewport=800 680 980 830, scale=0.66]{./figures/000102_detail_manip_gt.jpg}} \hspace{-1.5mm}
\subfigure[\scriptsize CAN($L_2$) \cite{chen2017fast}]{
\includegraphics*[viewport=800 680 980 830, scale=0.66]{./figures/000102_detail_manip_l2.jpg}} \hspace{-1.5mm}
\subfigure[\scriptsize CAN($L_2$+NIMA)]{
\includegraphics*[viewport=800 680 980 830, scale=0.66]{./figures/000102_detail_manip_l2_nima.jpg}} \hspace{0mm}
\end{center}
\vspace{-8 mm}
{\caption{Comparison of local detail enhancement operators in bright regions. In comparison to local tone mappers in \cite{Sylvain2011} and \cite{chen2017fast}, image details in dark and bright areas are better preserved in our results. \label{fig:bright}}}
\vspace{-4 mm}
\end{figure*}

\begin{figure*}[!t]
\vspace{-0 mm}
\begin{center}
\includegraphics*[scale=0.0733]{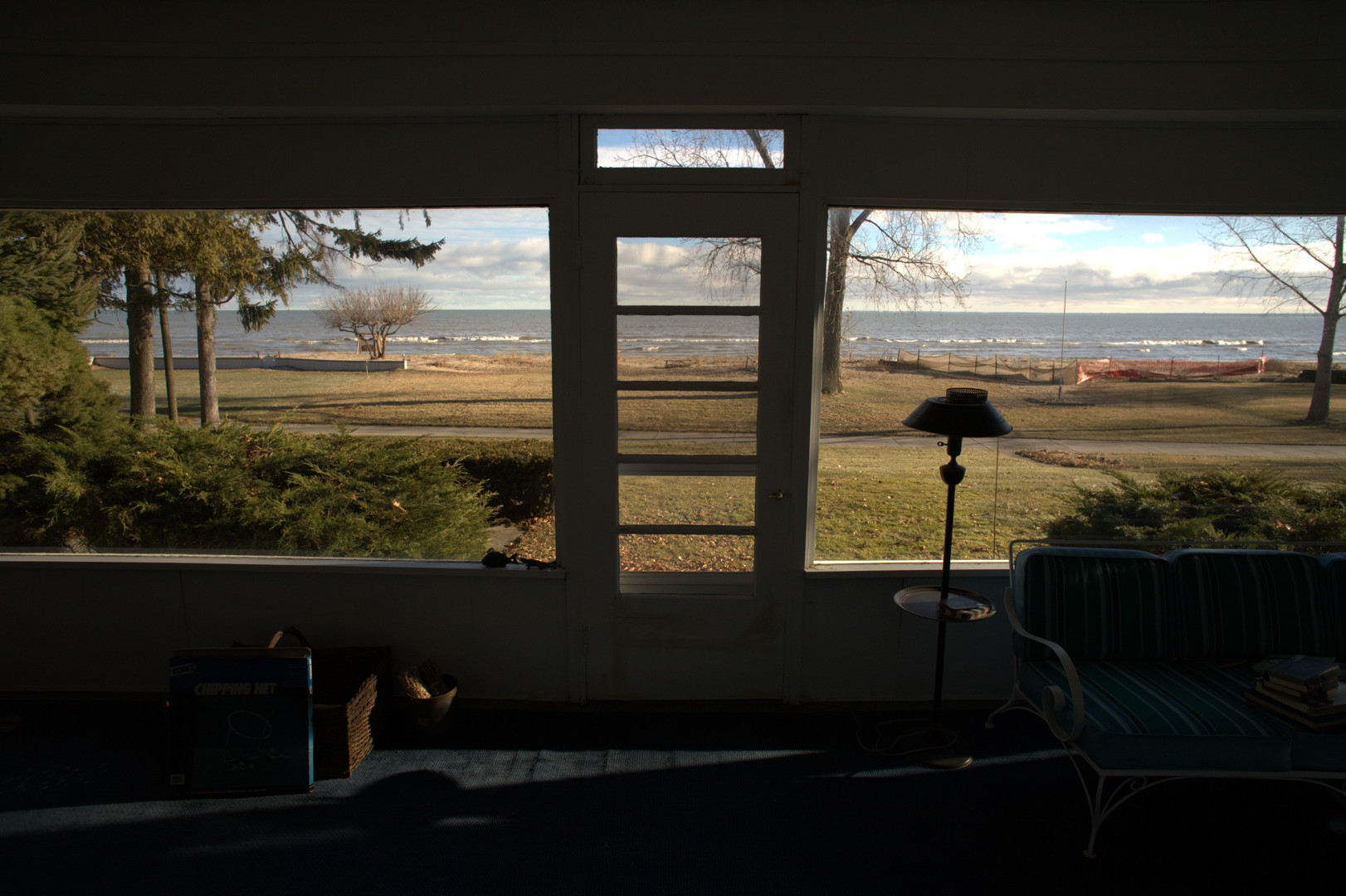}
\includegraphics*[scale=0.0733]{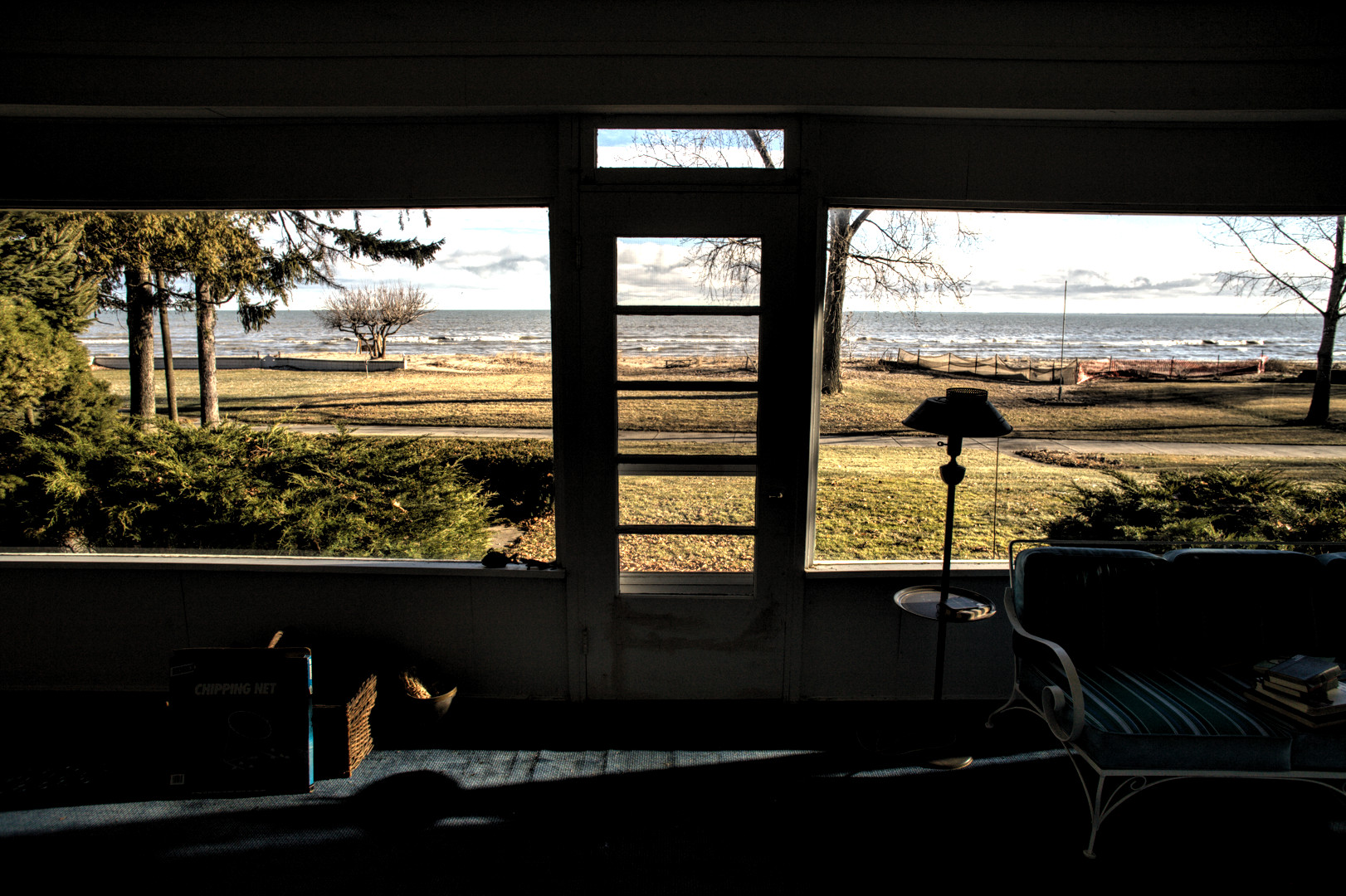}
\includegraphics*[scale=0.0733]{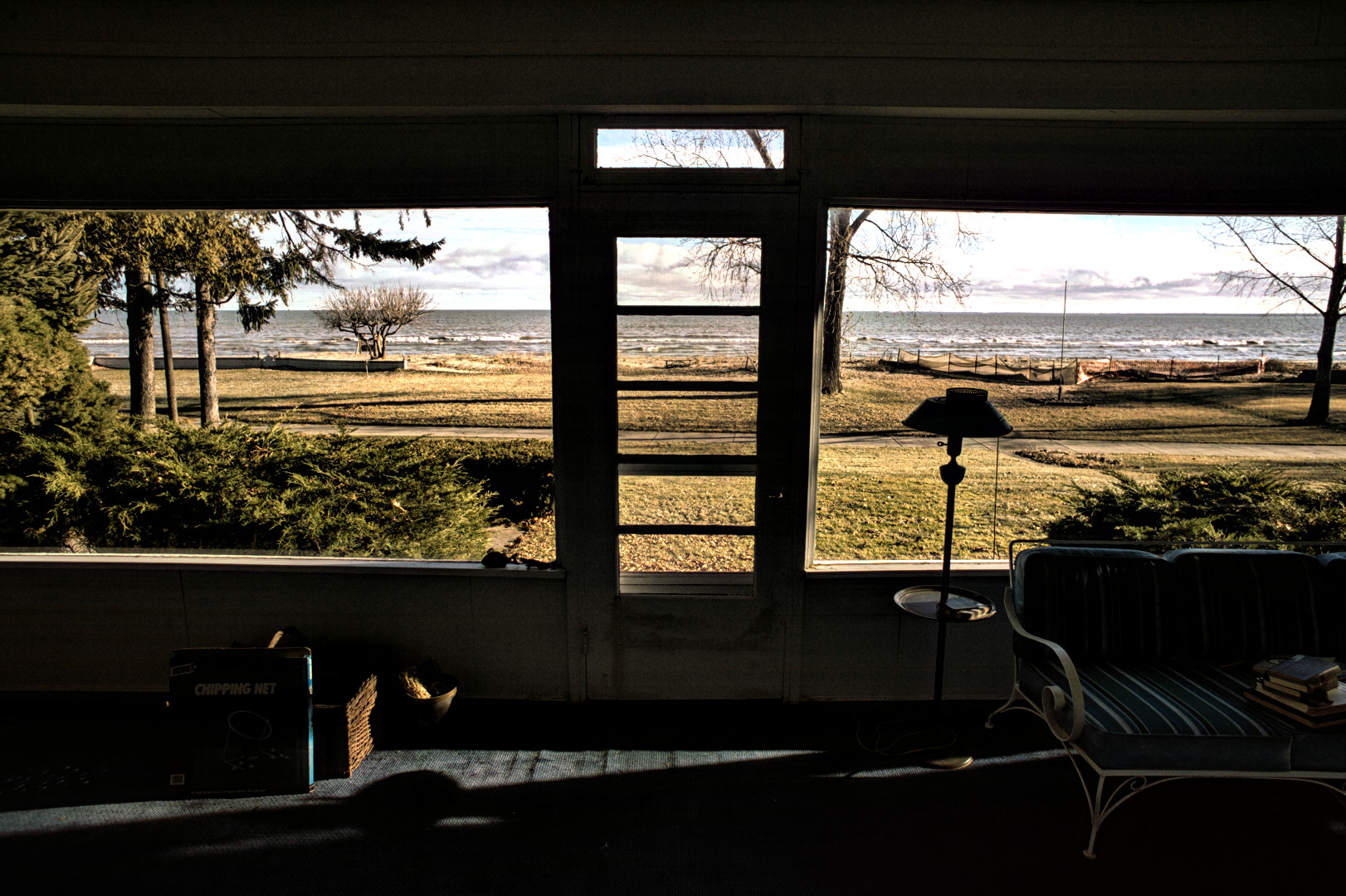}
\includegraphics*[scale=0.0733]{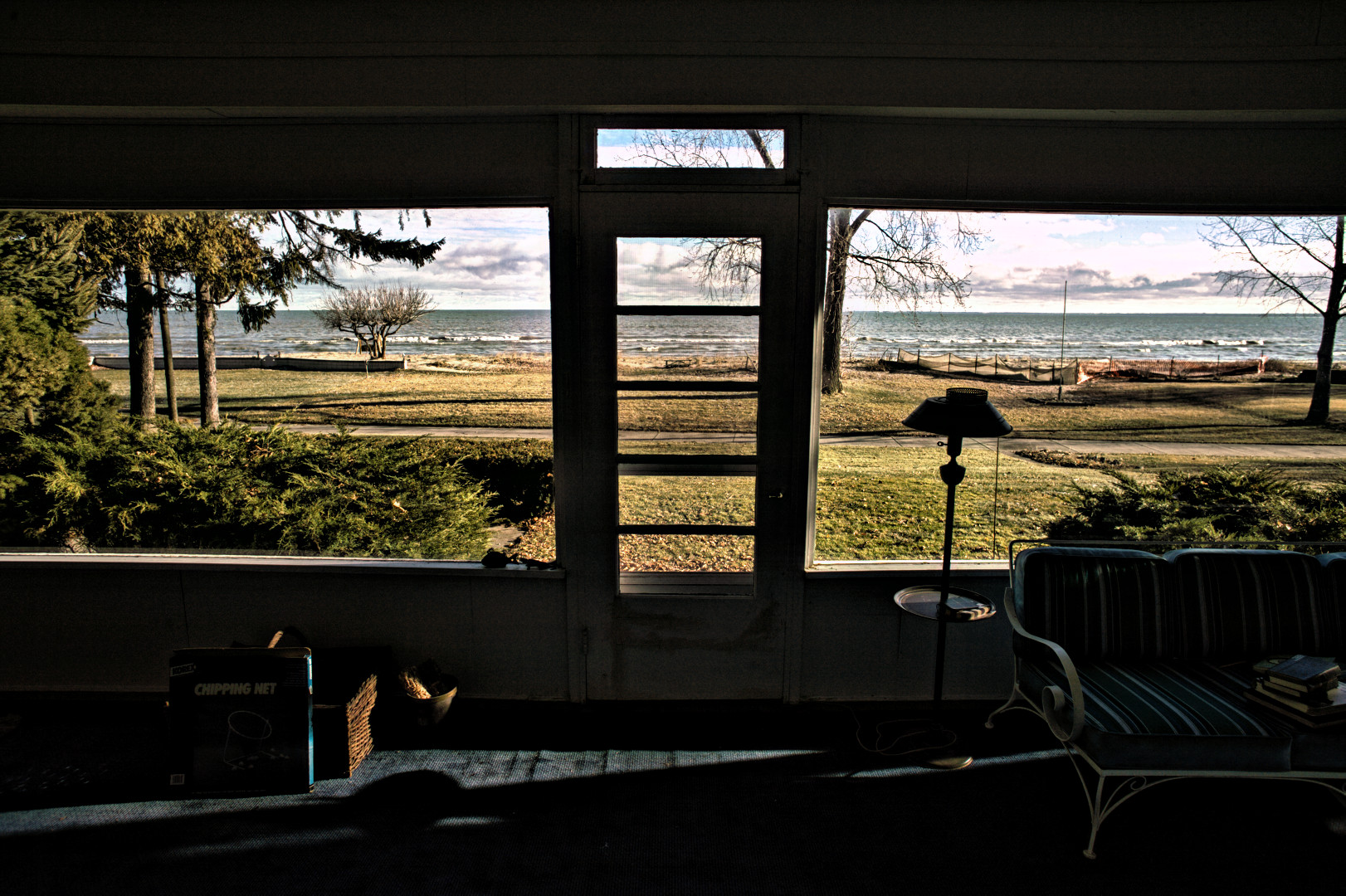}
\includegraphics*[viewport=480 710 650 830, scale=0.698]{./figures/000198.jpg}
\includegraphics*[viewport=480 710 650 830, scale=0.698]{./figures/000198_detail_manip_gt.jpg}
\includegraphics*[viewport=480 710 650 830, scale=0.698]{./figures/000198_detail_manip_l2.jpg}
\includegraphics*[viewport=480 710 650 830, scale=0.698]{./figures/000198_detail_manip_l2_nima.jpg}
\includegraphics*[viewport=1230 290 1400 400, scale=0.698]{./figures/000198.jpg}
\includegraphics*[viewport=1230 290 1400 400, scale=0.698]{./figures/000198_detail_manip_gt.jpg}
\includegraphics*[viewport=1230 290 1400 400, scale=0.698]{./figures/000198_detail_manip_l2.jpg}
\includegraphics*[viewport=1230 290 1400 400, scale=0.698]{./figures/000198_detail_manip_l2_nima.jpg}
\includegraphics*[scale=0.0732]{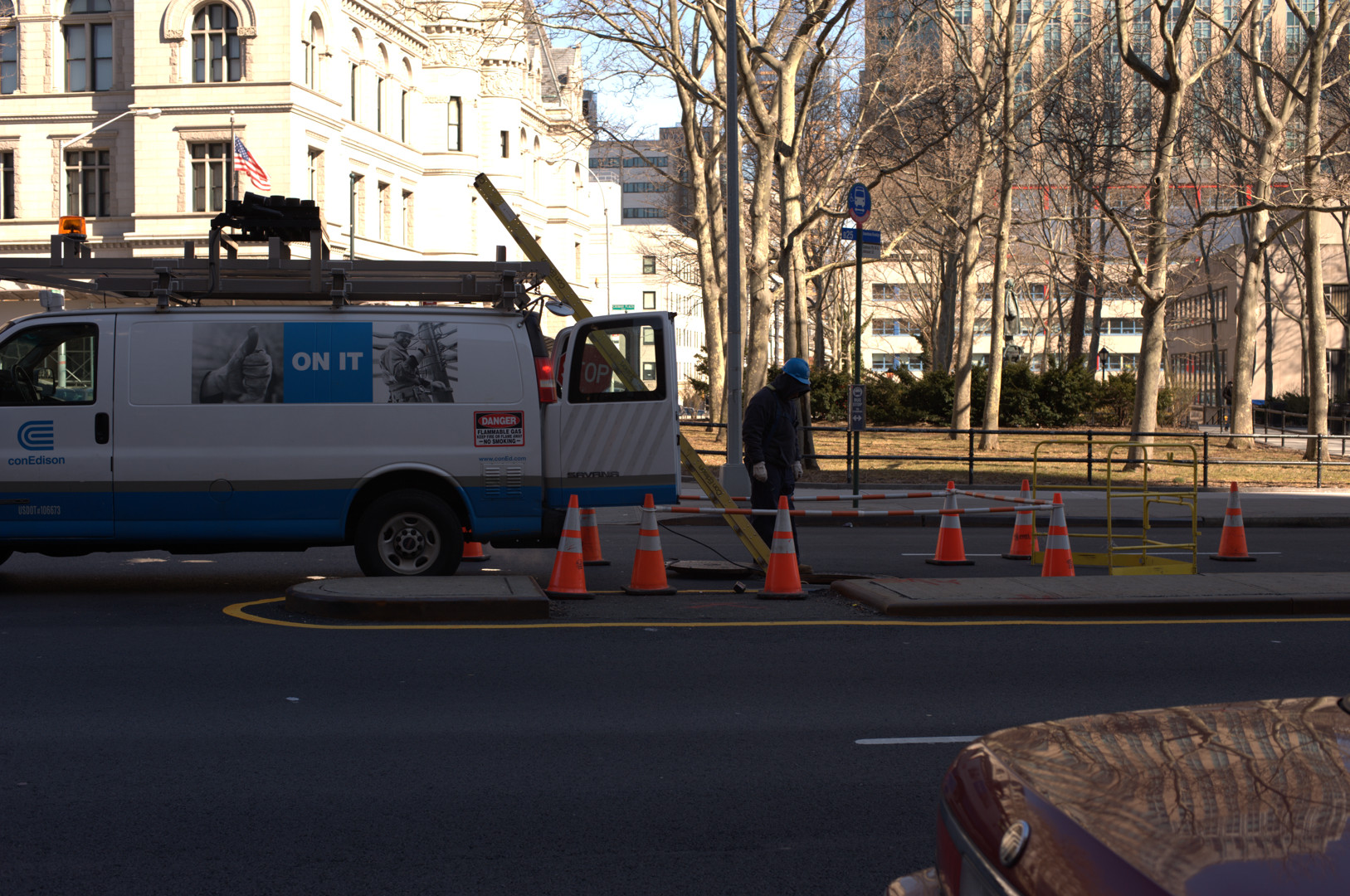}
\includegraphics*[scale=0.0732]{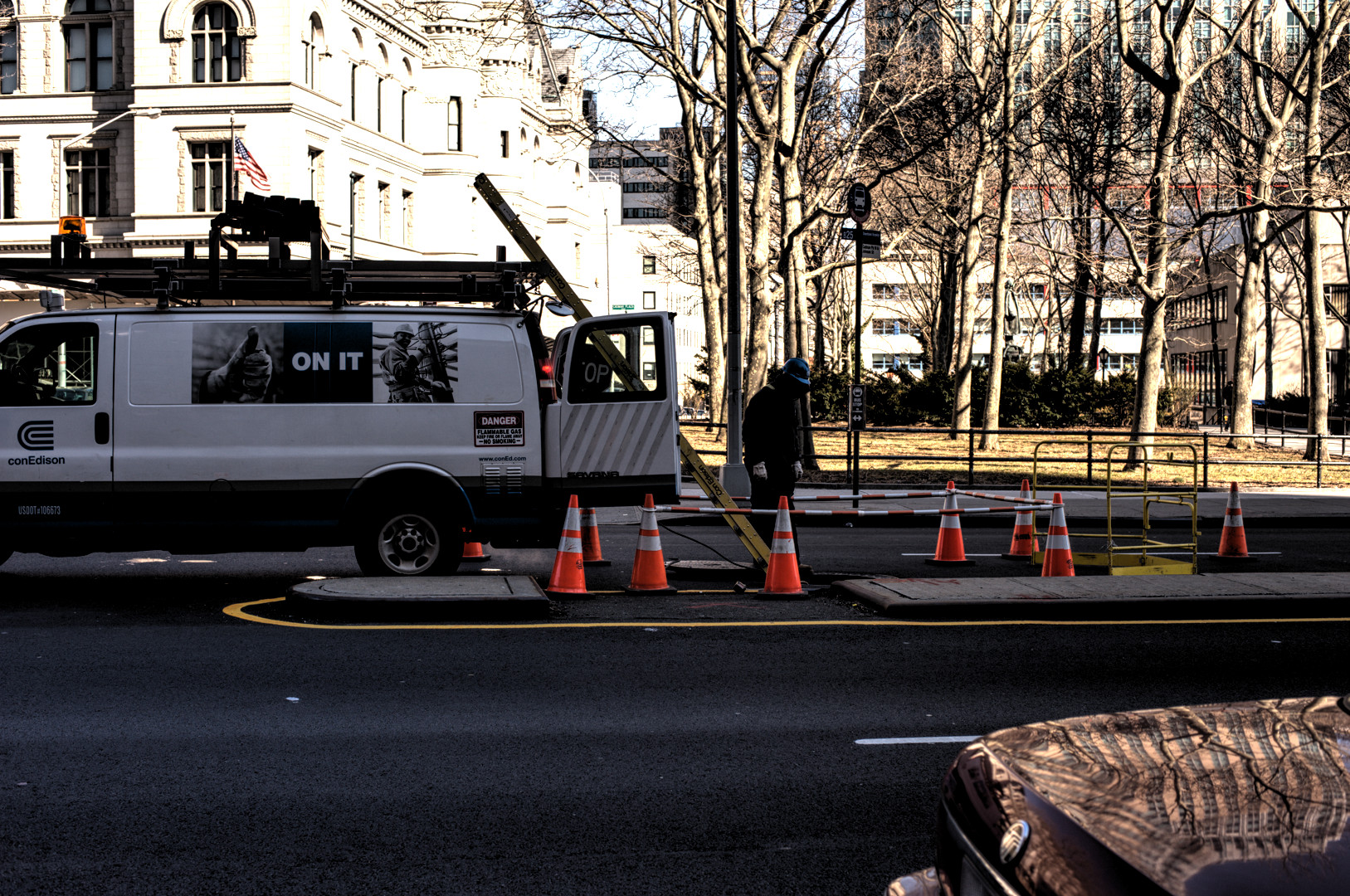}
\includegraphics*[scale=0.0732]{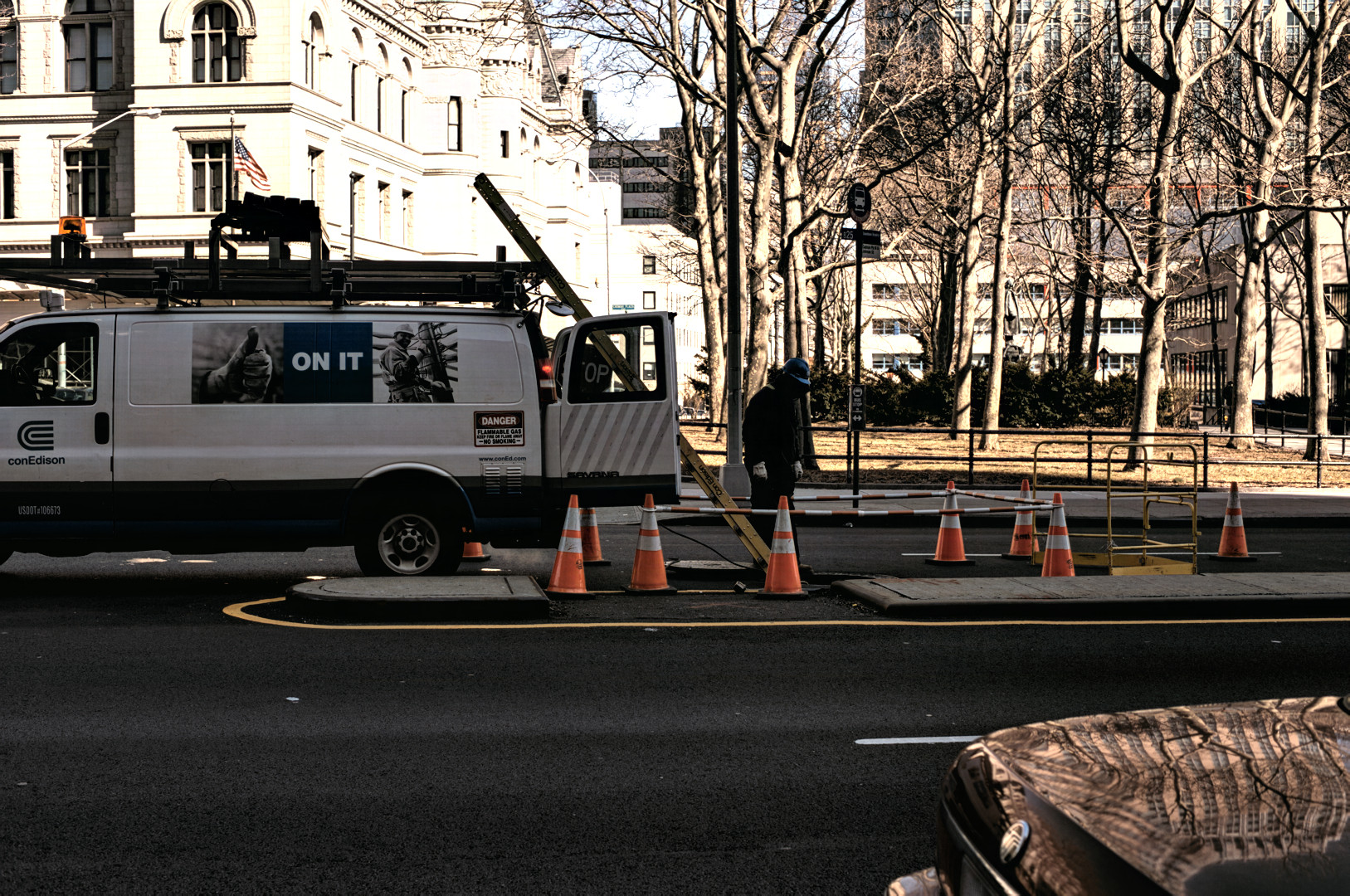}
\includegraphics*[scale=0.0732]{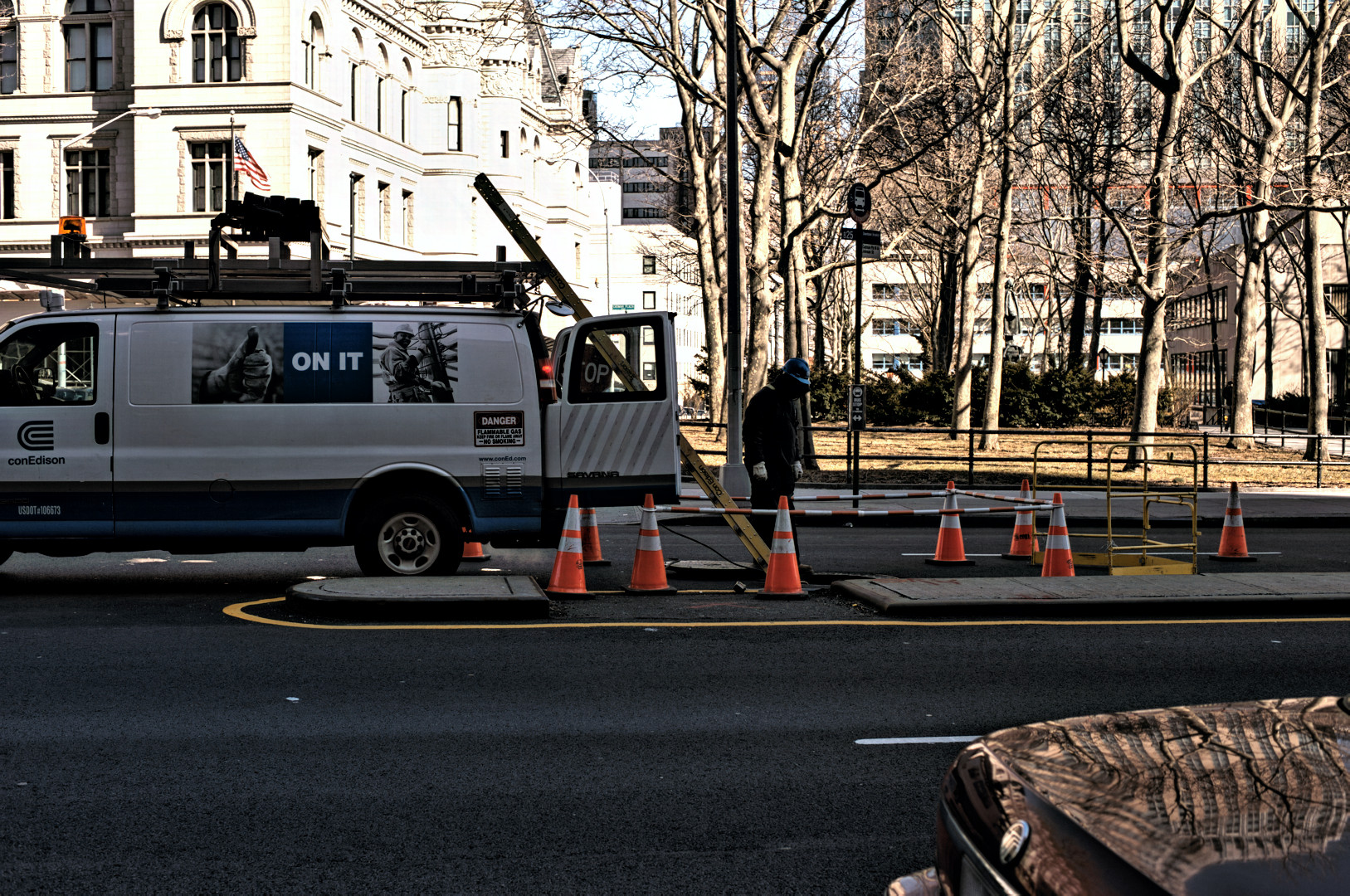}
\includegraphics*[viewport=500 400 680 500, scale=0.66]{./figures/000020.jpg}
\includegraphics*[viewport=500 400 680 500, scale=0.66]{./figures/000020_detail_manip_gt.jpg}
\includegraphics*[viewport=500 400 680 500, scale=0.66]{./figures/000020_detail_manip_l2.jpg}
\includegraphics*[viewport=500 400 680 500, scale=0.66]{./figures/000020_detail_manip_l2_nima.jpg}
\includegraphics*[scale=0.0733]{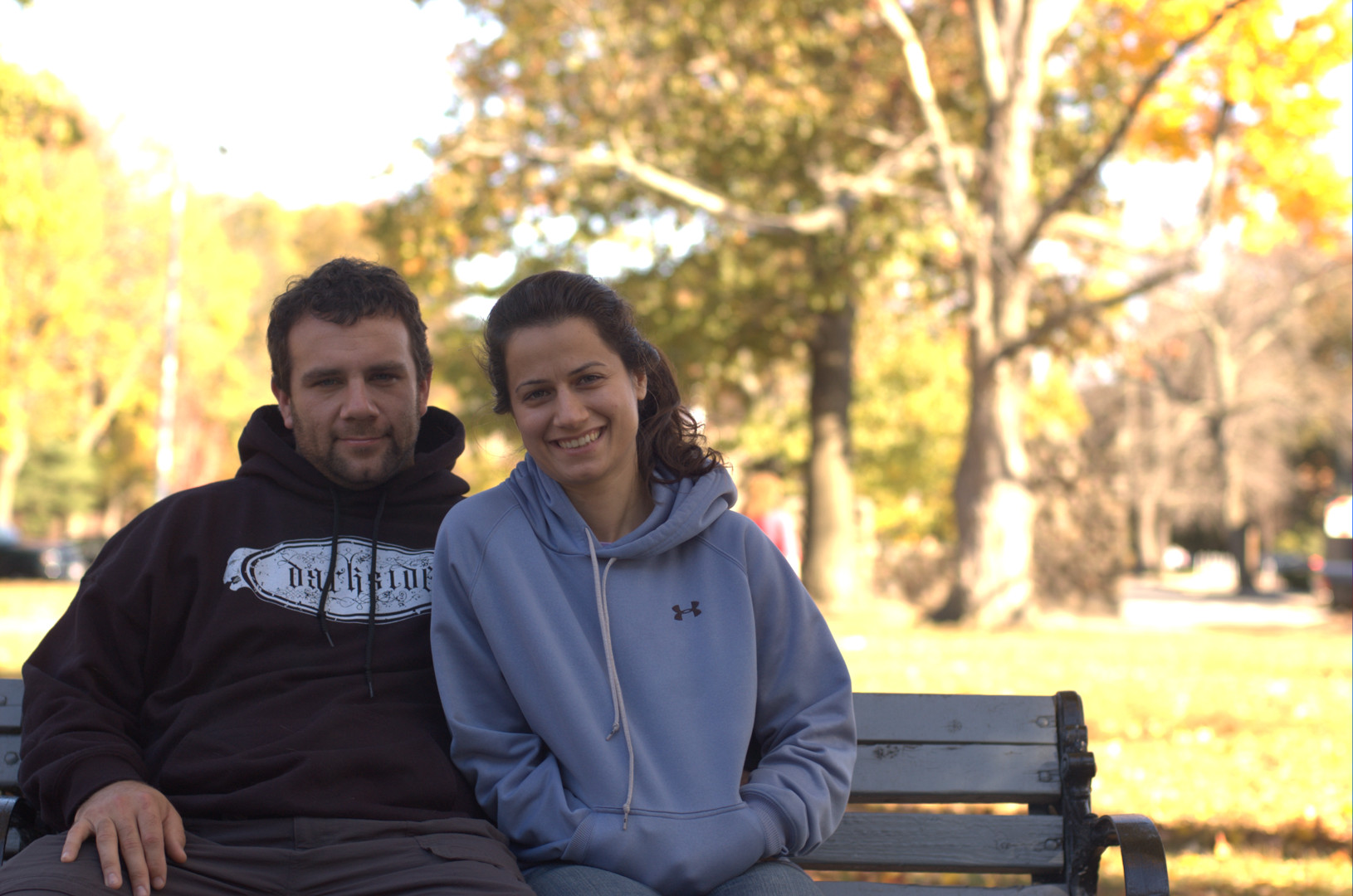}
\includegraphics*[scale=0.0733]{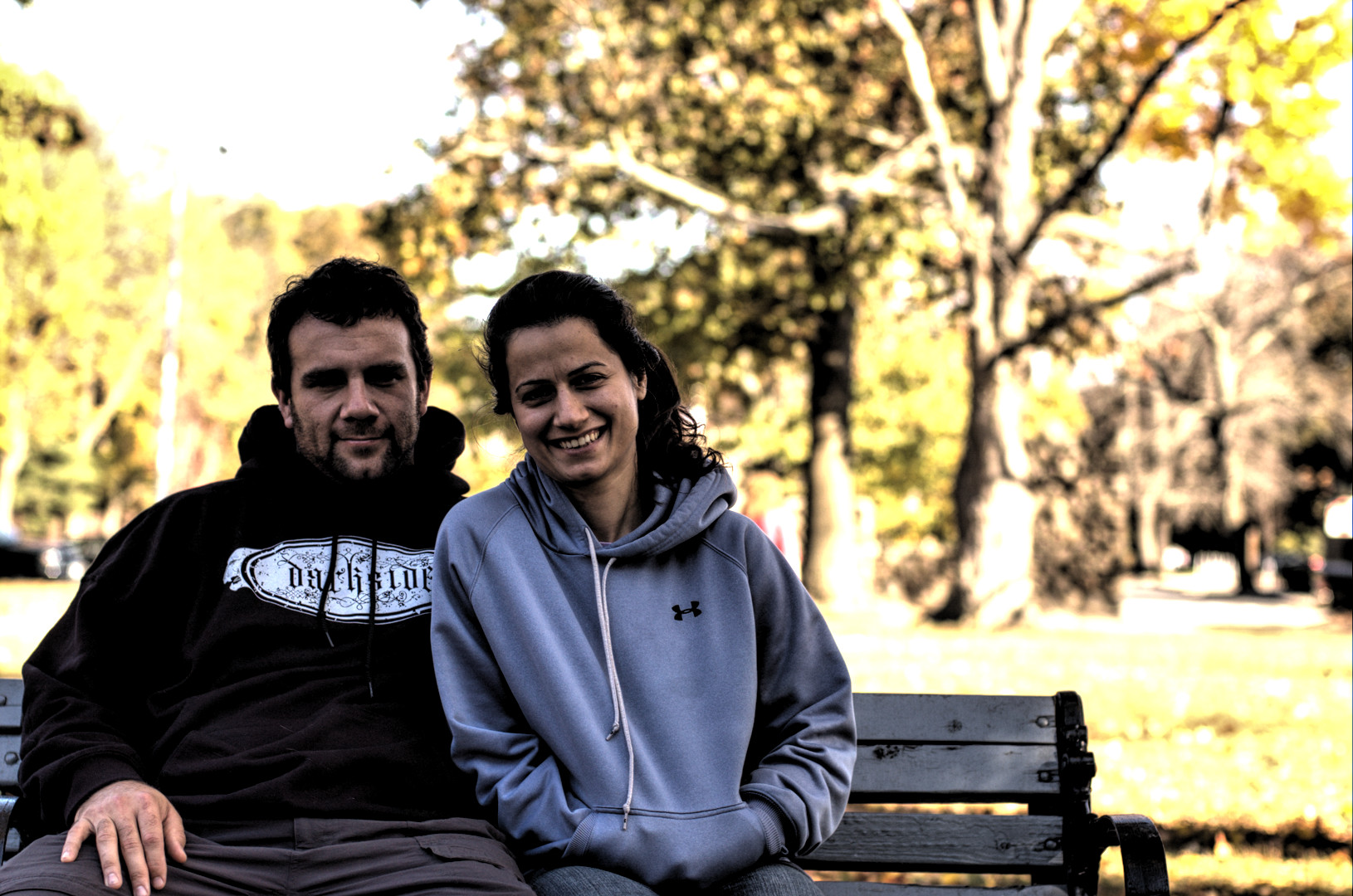}
\includegraphics*[scale=0.0733]{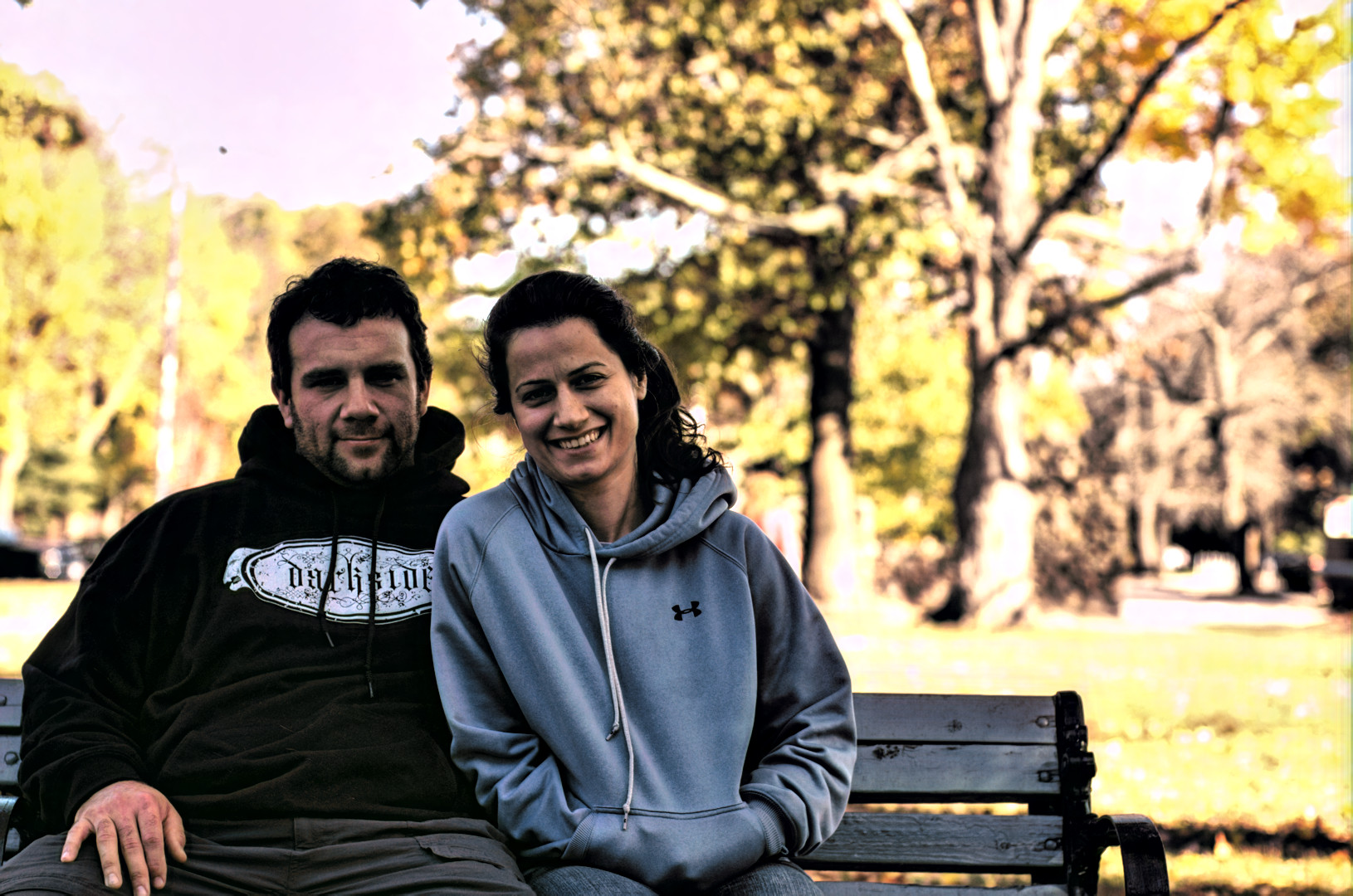}
\includegraphics*[scale=0.0733]{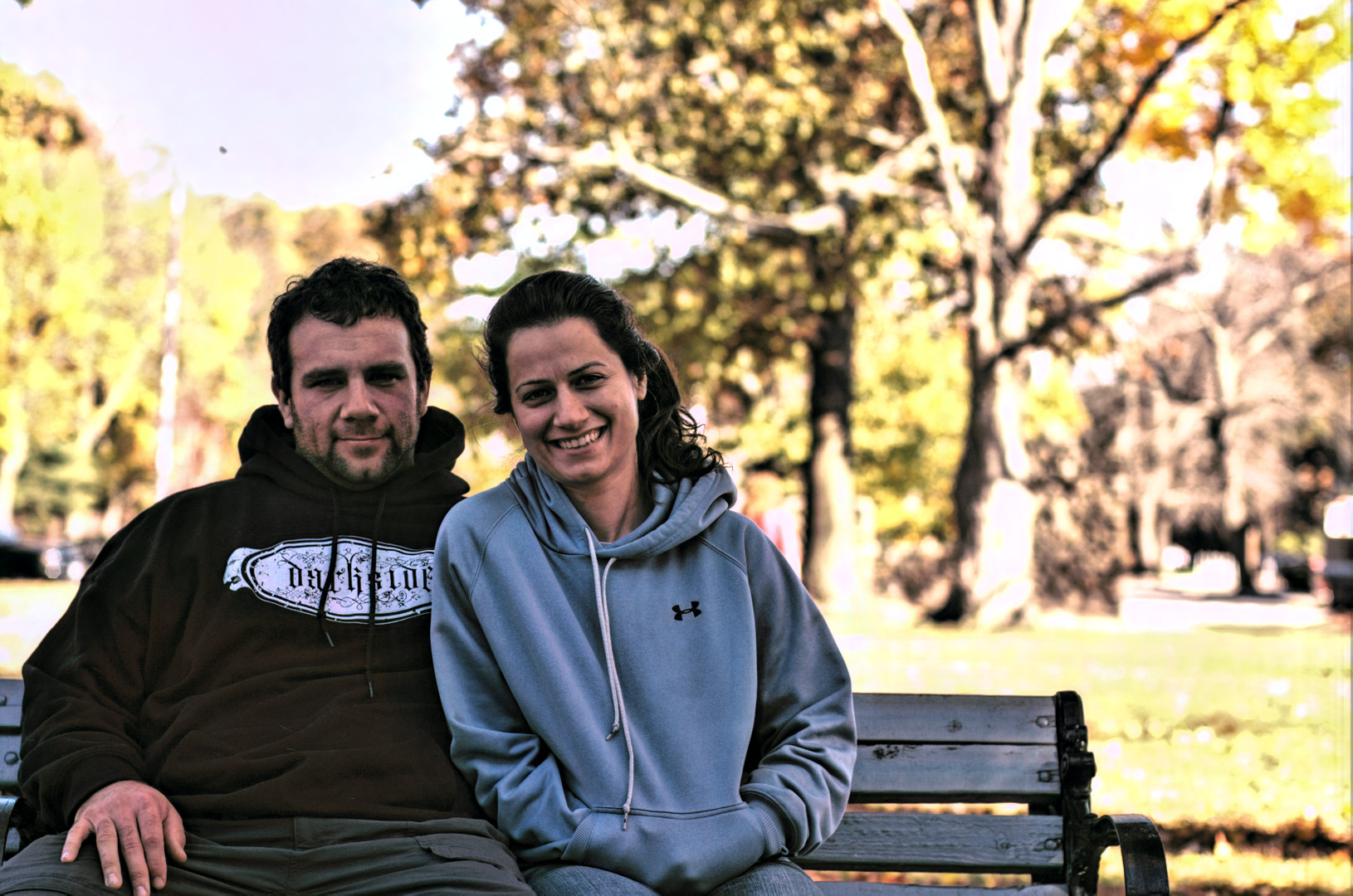}
\\[-1.7mm]
\subfigure[\scriptsize Input]{
\includegraphics*[viewport=600 510 780 700, scale=0.6665]{./figures/000209.jpg}} \hspace{-2mm}
\subfigure[\scriptsize Reference \cite{Sylvain2011}]{
\includegraphics*[viewport=600 510 780 700, scale=0.6665]{./figures/000209_detail_manip_gt.jpg}} \hspace{-2mm}
\subfigure[\scriptsize CAN($L_2$) \cite{chen2017fast}]{
\includegraphics*[viewport=600 510 780 700, scale=0.6665]{./figures/000209_detail_manip_l2.jpg}} \hspace{-2mm}
\subfigure[\scriptsize CAN($L_2$+NIMA)]{
\includegraphics*[viewport=600 510 780 700, scale=0.6665]{./figures/000209_detail_manip_l2_nima.jpg}} \hspace{-2mm}
\end{center}
\vspace{-3 mm}
{\caption{Comparison of various local detail enhancement operators on high contrast photos. In comparison to local tone mappers in \cite{Sylvain2011} and \cite{chen2017fast}, image details in dark and bright areas are better preserved in our results. \label{fig:contrast}}}
\vspace{-0 mm}
\end{figure*}

\begin{figure*}[!t]
\vspace{-0 mm}
\begin{center}
\includegraphics*[scale=0.0733]{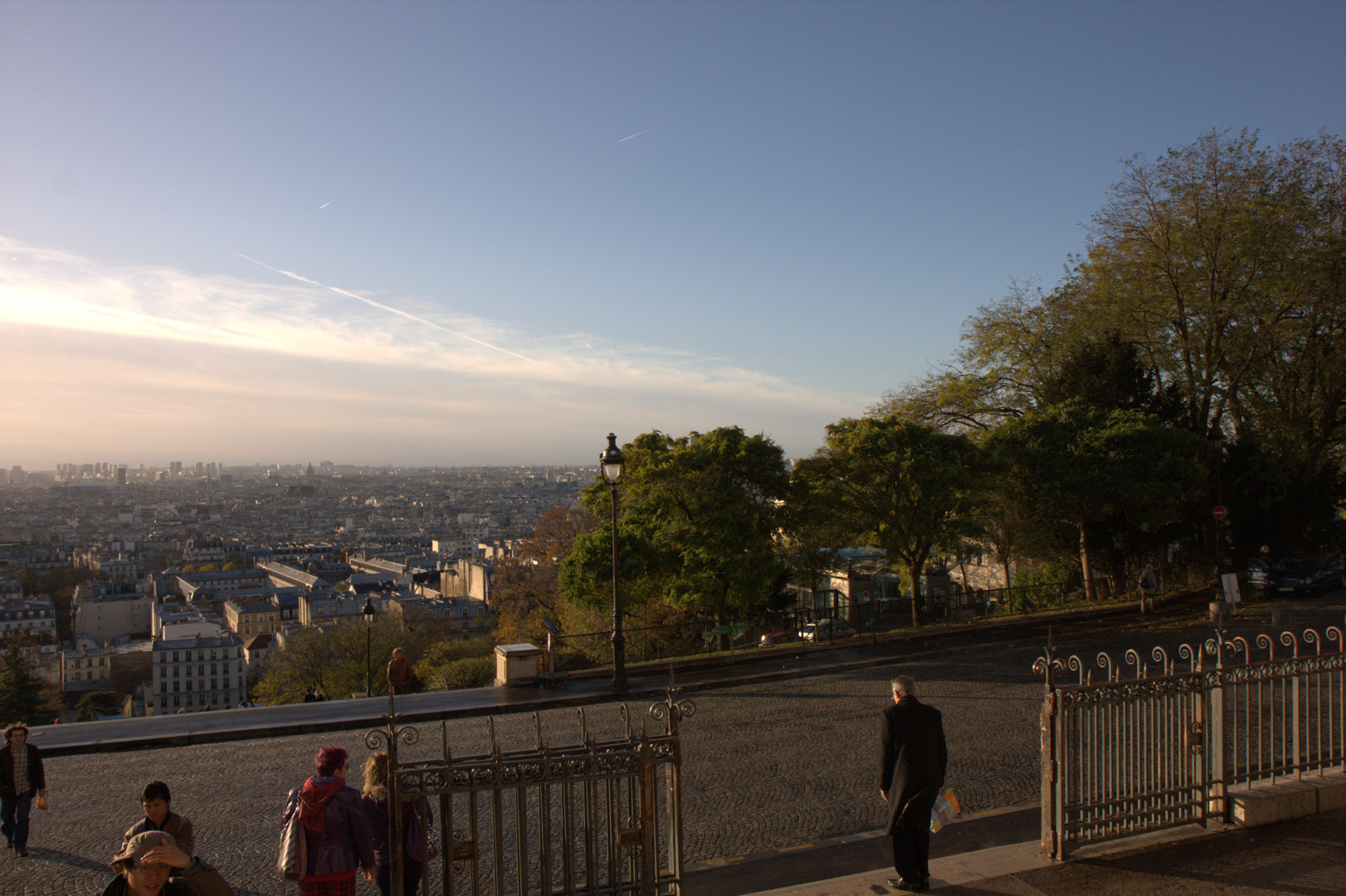}
\includegraphics*[scale=0.0733]{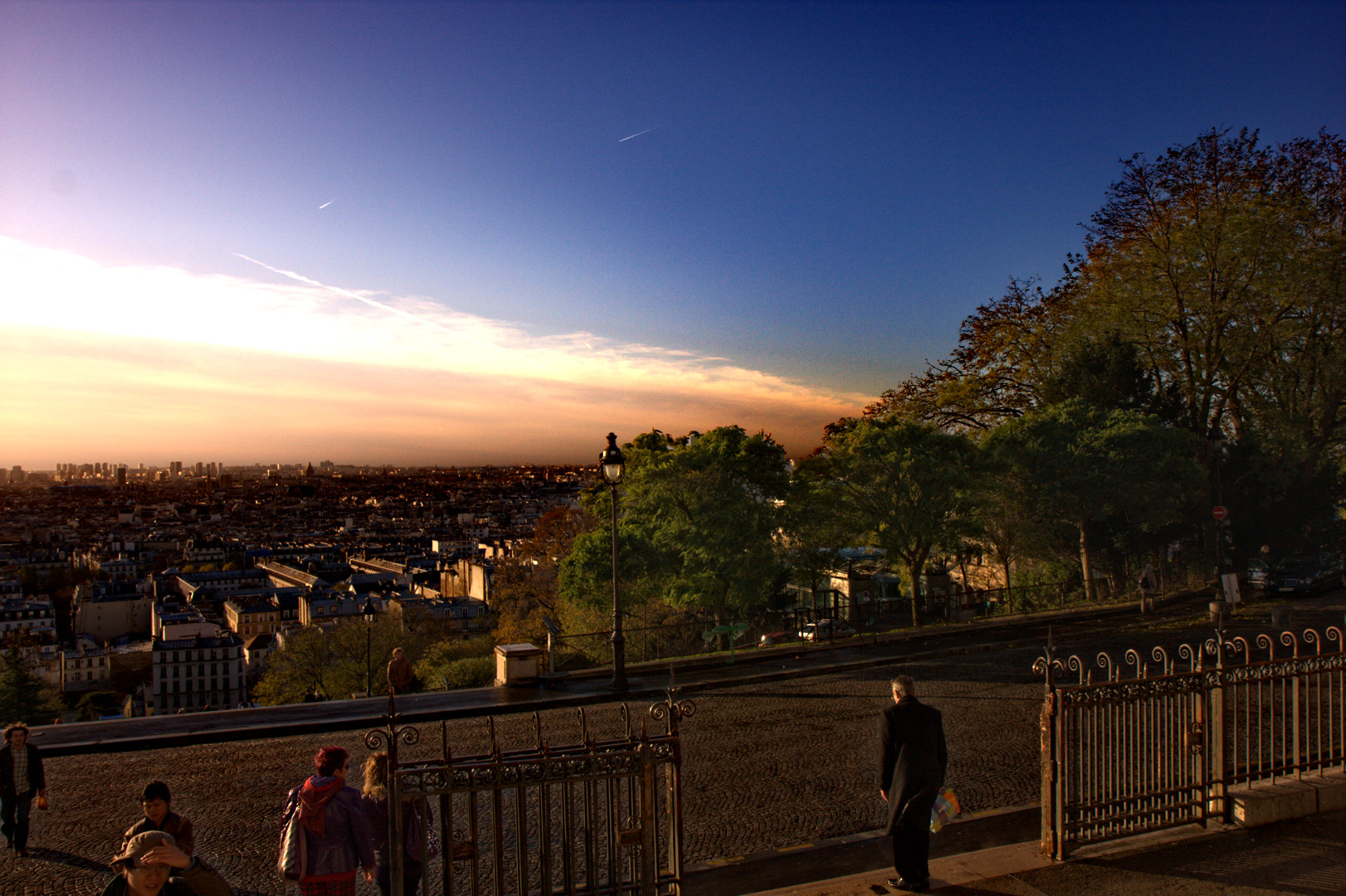}
\includegraphics*[scale=0.0733]{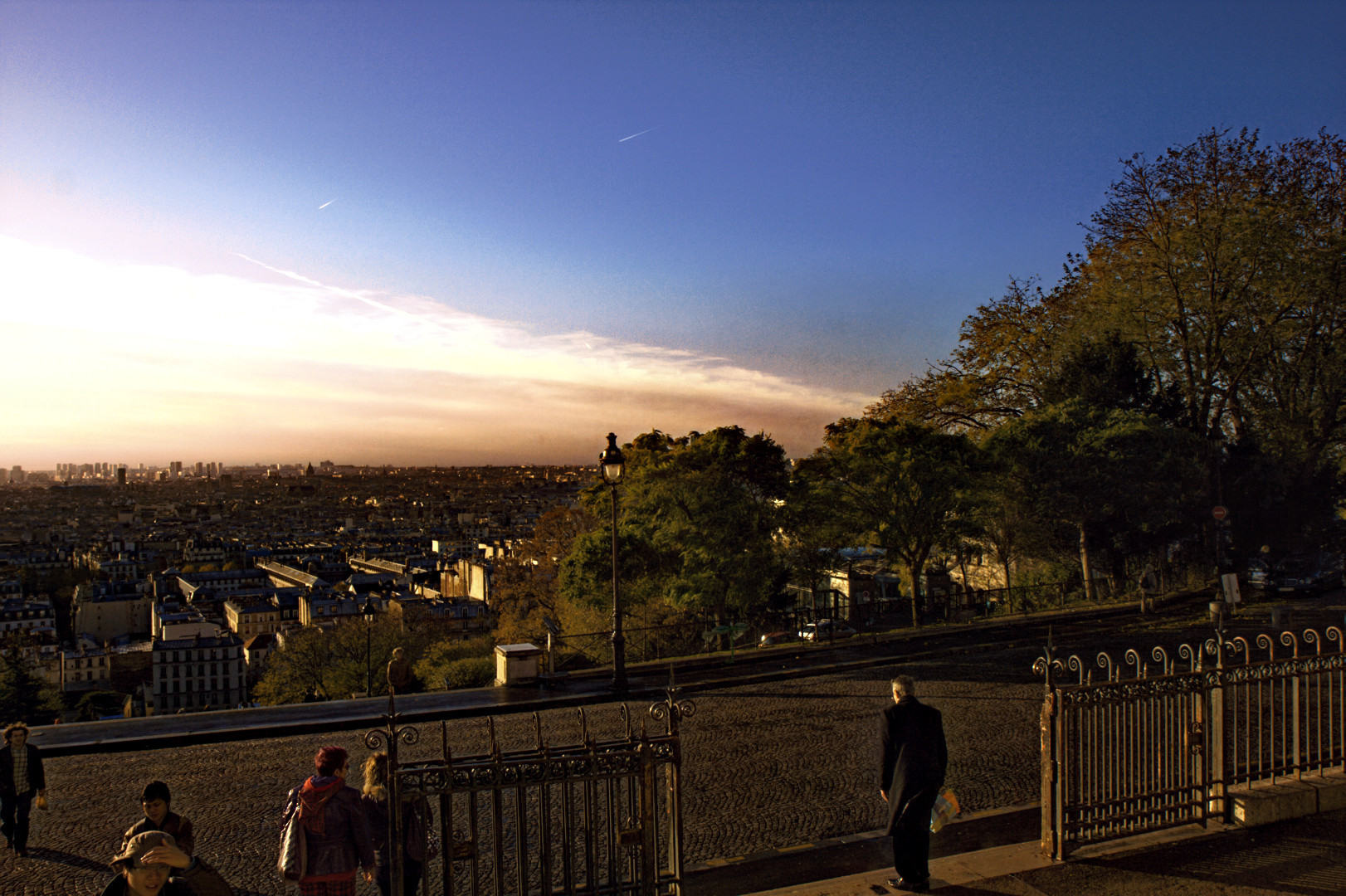}
\includegraphics*[scale=0.0733]{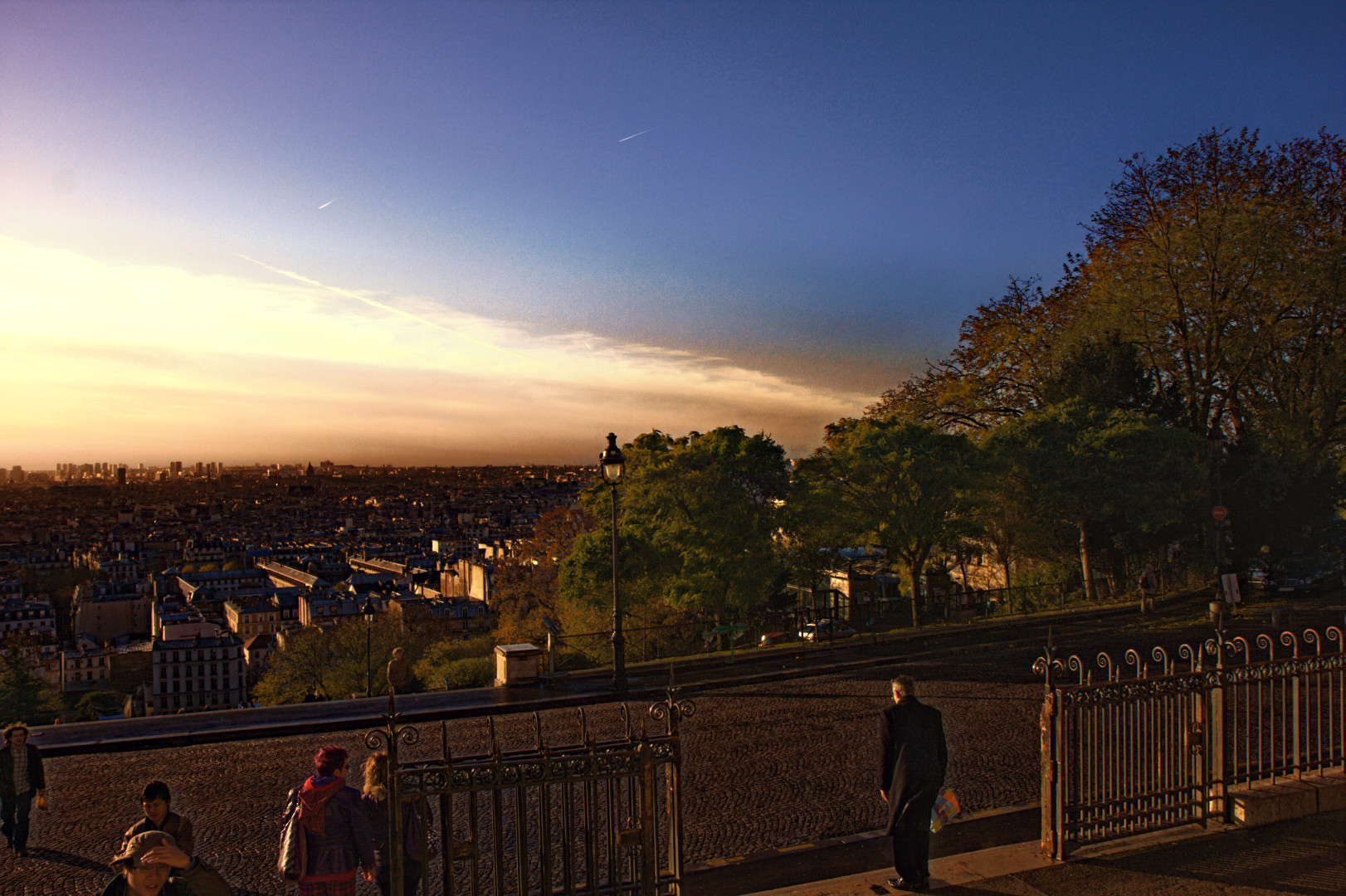}
\includegraphics*[viewport=300 340 480 480, scale=0.661]{./figures/000060.jpg}
\includegraphics*[viewport=300 340 480 480, scale=0.661]{./figures/000060_dehazing_gt.jpg}
\includegraphics*[viewport=300 340 480 480, scale=0.661]{./figures/000060_dehazing_l2.jpg}
\includegraphics*[viewport=300 340 480 480, scale=0.661]{./figures/000060_dehazing_l2_nima.jpg}
\includegraphics*[scale=0.0733]{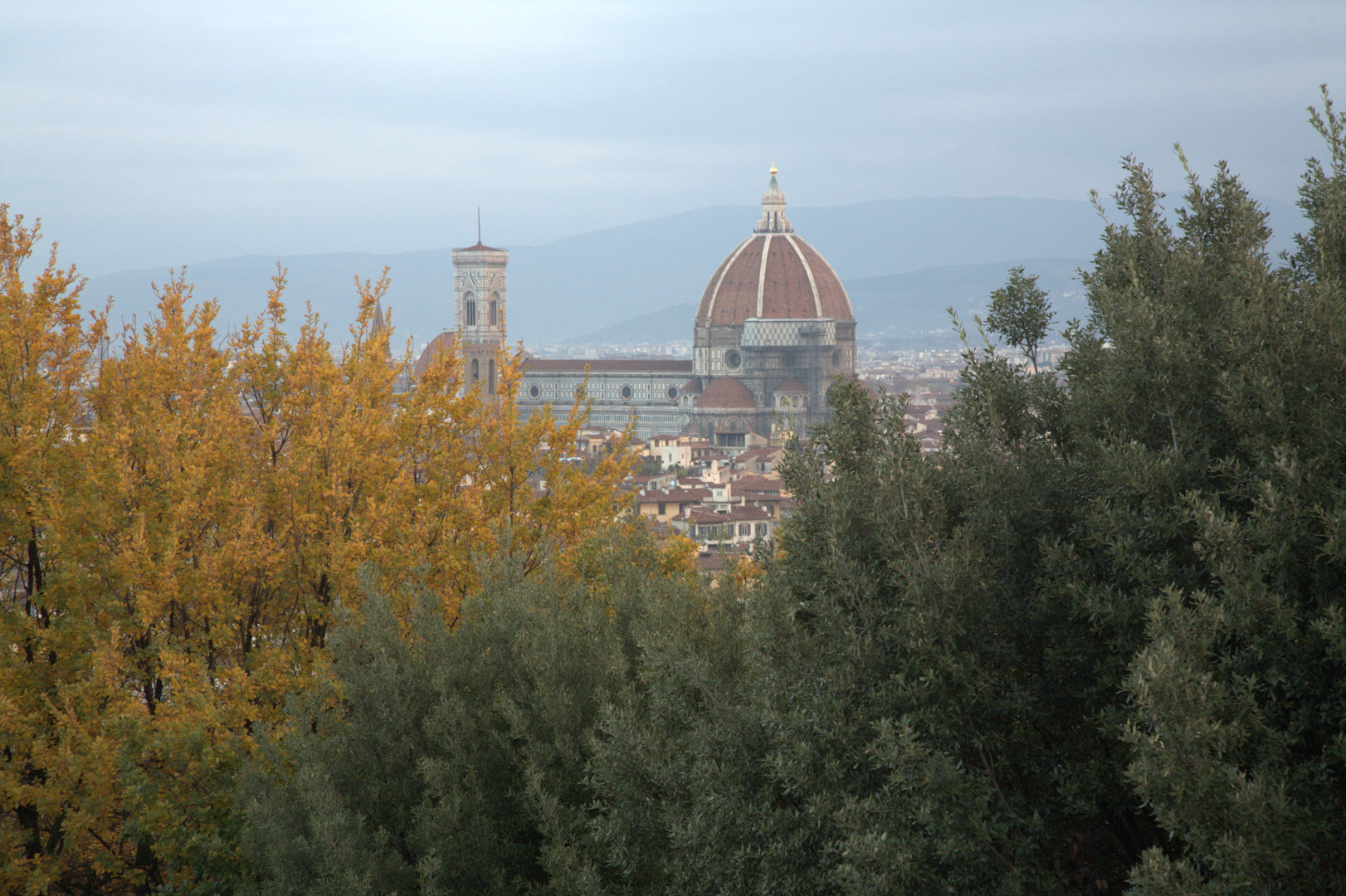}
\includegraphics*[scale=0.0733]{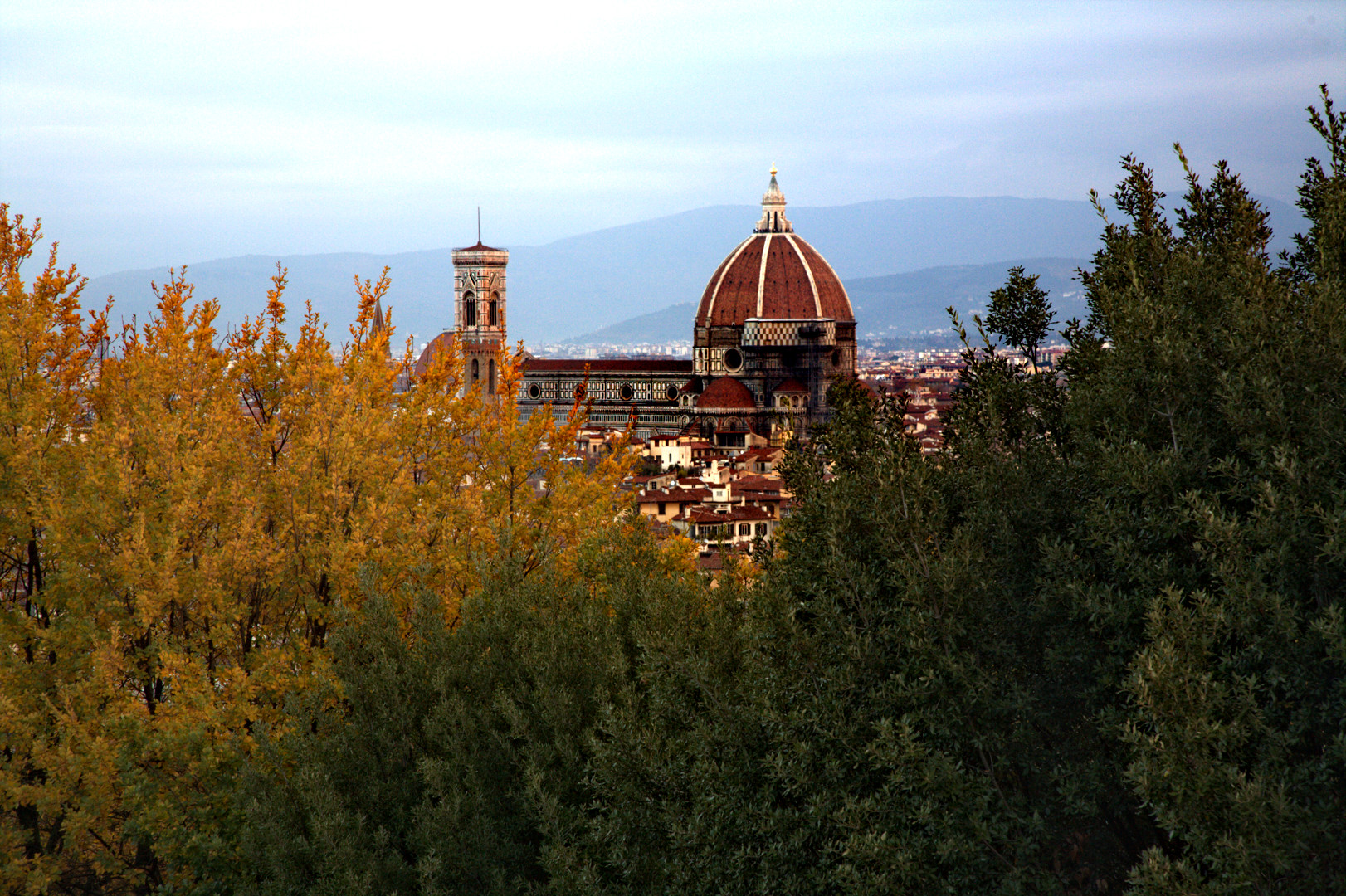}
\includegraphics*[scale=0.0733]{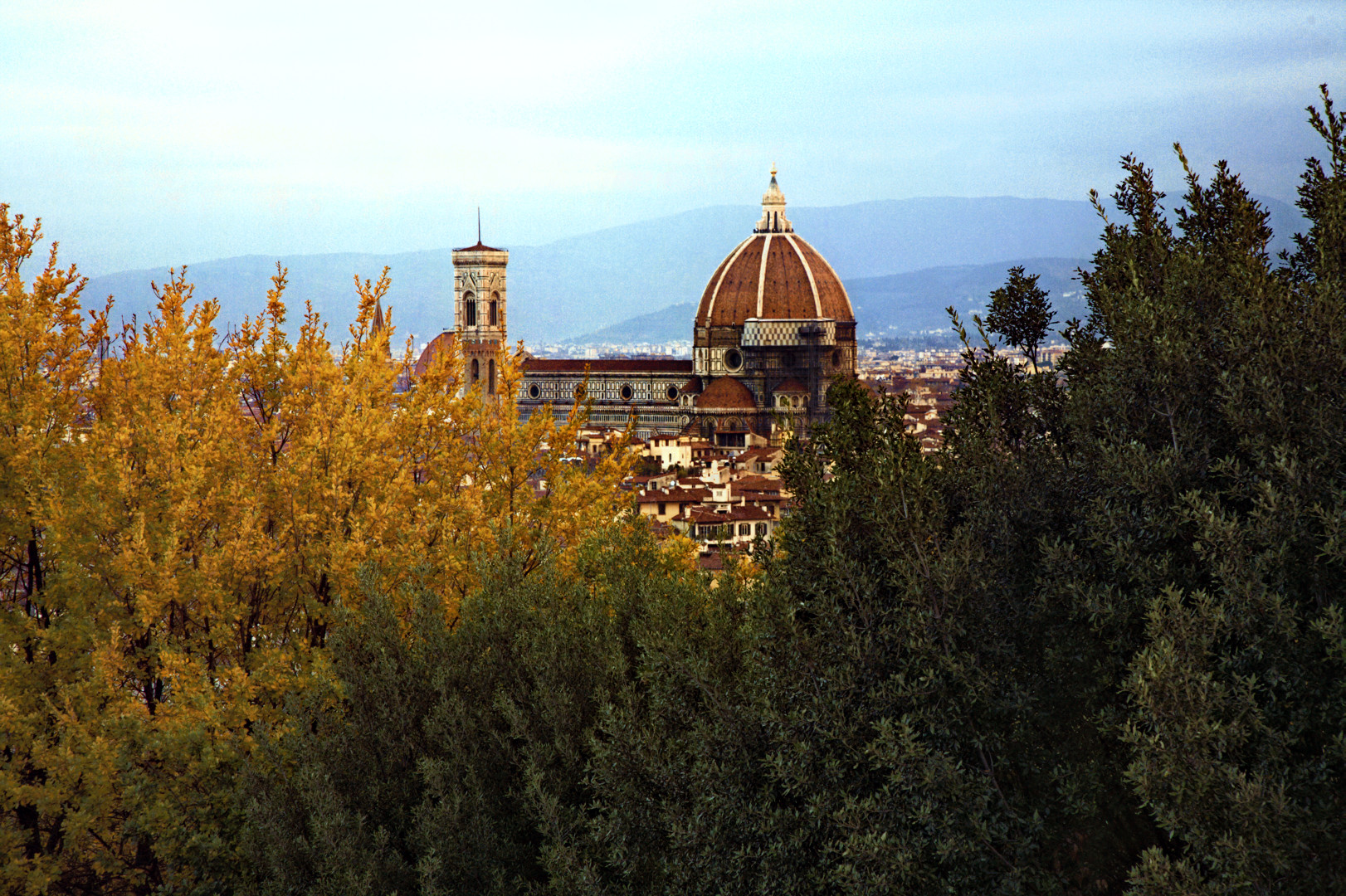}
\includegraphics*[scale=0.0733]{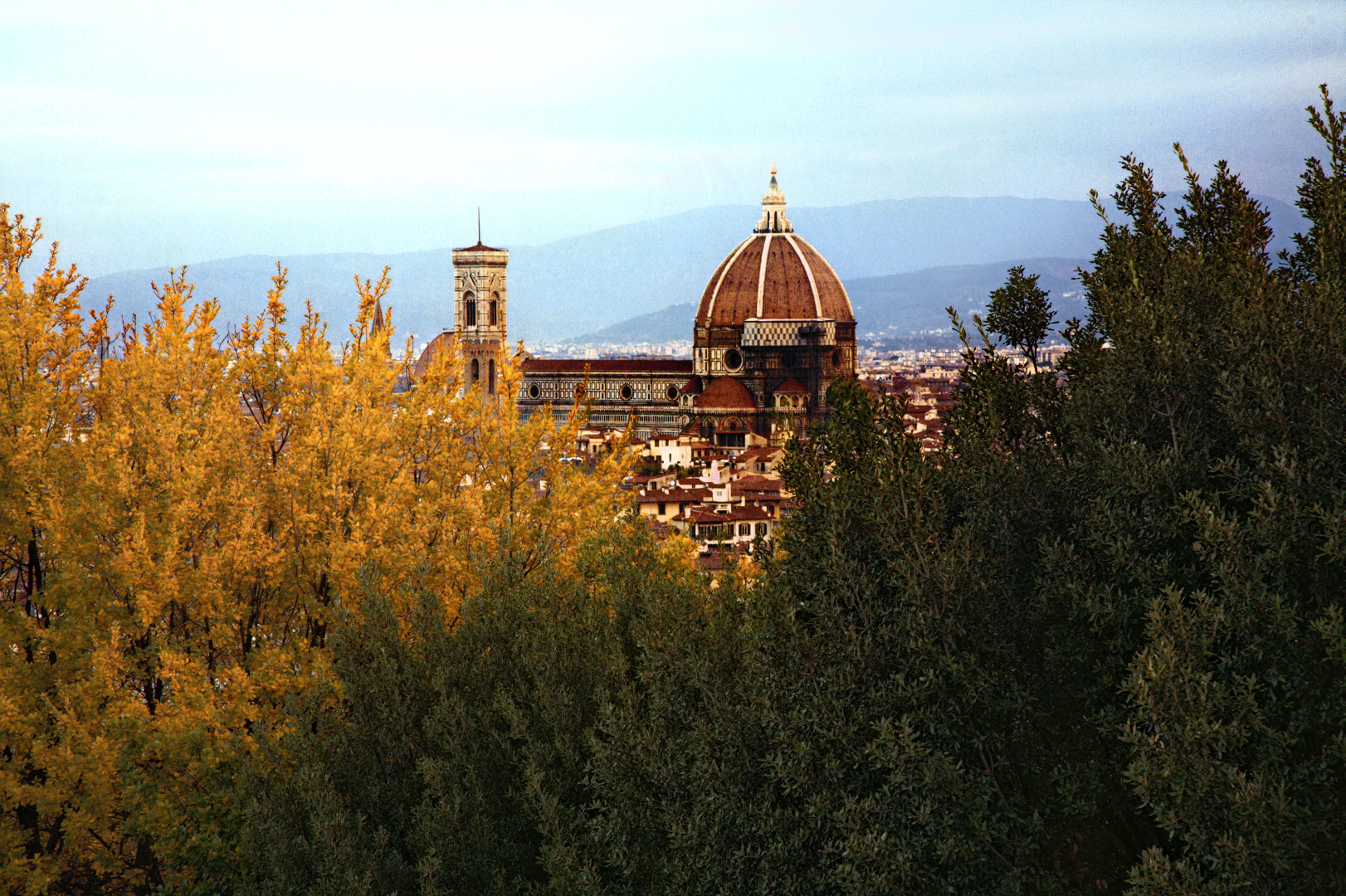}
\\[-1.7mm]
\subfigure[\scriptsize Input]{
\includegraphics*[viewport=800 530 980 680, scale=0.659]{./figures/000142.jpg}} \hspace{-1.5mm}
\subfigure[\scriptsize Reference\cite{NonLocalImageDehazing}]{
\includegraphics*[viewport=800 530 980 680, scale=0.659]{./figures/000142_dehazing_gt.jpg}} \hspace{-1.5mm}
\subfigure[\scriptsize CAN($L_2$) \cite{chen2017fast}]{
\includegraphics*[viewport=800 530 980 680, scale=0.659]{./figures/000142_dehazing_l2.jpg}} \hspace{-1.75mm}
\subfigure[\scriptsize CAN($L_2$+NIMA)]{
\includegraphics*[viewport=800 530 980 680, scale=0.659]{./figures/000142_dehazing_l2_nima.jpg}} \hspace{-1.0mm}
\end{center}
\vspace{-3 mm}
{\caption{Comparison of image dehazing operators. In comparison to nonlocal dehazing in \cite{NonLocalImageDehazing} and \cite{chen2017fast}, image color palette and local tone of our results  are superior. \label{fig:dehazing}}}
\vspace{-0 mm}
\end{figure*}

Chen et al. \cite{chen2017fast} show that training CAN with $L_2$ loss leads to approximation of tone enhancement operator \cite{Sylvain2011} and nonlocal image dehazing \cite{NonLocalImageDehazing}. However, our visual inspections indicate that tone mapping with CAN may result in poor detail preservation in dark areas (Fig.~\ref{fig:dark}) and blown-out highlights (Fig.~\ref{fig:bright}). Also, stretching contrast of photos may cause washed out or faded colors (Fig.~\ref{fig:contrast}). Similar issue can be observed in the approximation of nonlocal image dehazing (Fig.~\ref{fig:dehazing}). Comparing the tone and detail enhancement results in Fig.~\ref{fig:dark}, Fig.~\ref{fig:bright}, and Fig.~\ref{fig:contrast} indicate that our perceptual loss improves upon results from \cite{Sylvain2011} and \cite{chen2017fast}. Interestingly, despite the lost details in the reference image \cite{Sylvain2011}, the perceptual measure allows preserving and enhancing details in dark {\em and} bright areas. Our results on image dehazing in Fig.~\ref{fig:dehazing} present better color saturation compared to training with only $L_2$ loss in \cite{chen2017fast}.

\begin{figure}[!t]
\vspace{-0 mm}
\begin{center}
\includegraphics*[viewport=20 20 450 340, scale=0.5]{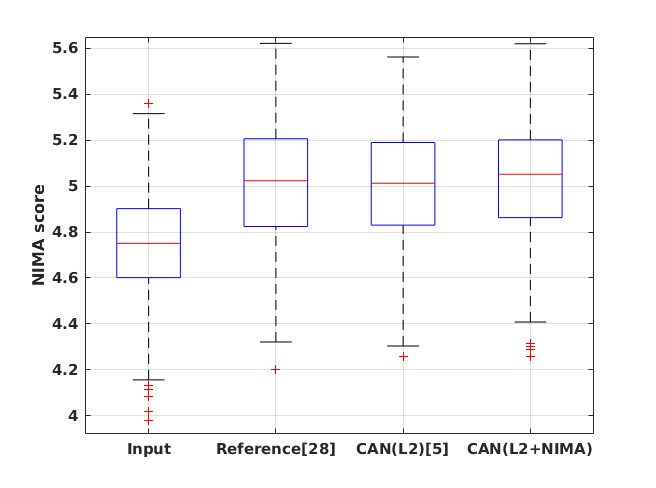}
\end{center}
\vspace{-4 mm}
{\caption{NIMA(Inception-v2) scores for input \cite{fivek}, reference \cite{Sylvain2011}, CAN($L_2$) \cite{chen2017fast}, and CAN($L_2$+NIMA).\label{fig:scores}}}
\vspace{-4 mm}
\end{figure}

NIMA score statistics associated with all FiveK test images are reported in Fig.~\ref{fig:scores}. Based on these scores, all detail enhancement methods are improving upon the input image; however, our perceptual loss shows higher average score, and lower variance about the mean.

\begin{table}[!t]
\begin{center}
\caption{Comparison of the proposed quality assessment method with various architectures. Numbers are reported for applying NIMA models on images of size $480\times640\times3$.}
\scalebox{0.9}{
\begin{tabular}{@{} *4l @{}}    \toprule
 &  NIMA & NIMA&  NIMA  \\
\emph{Model} & (MobileNet) & (Inception-v2) &  (VGG16)  \\\midrule
\emph{Million Parameters}   & 3.22 & 10.16 & 134.30  \\
\emph{Billion Flops}  &  6.97 & 23.80 & 218.43  \\\bottomrule
 \hline
\end{tabular}}
\label{tab:time_comp}
\end{center}
\vspace{-4 mm}
\end{table}

\section{Conclusions}

In this work a perceptual loss for image enhancement is introduced. This loss is built on a no-reference quality predictor trained on images annotated by human raters. Consequently, human visual preferences are encoded in our loss, and can be effectively used for guiding image enhancement algorithms. The proposed loss is a differentiable CNN, and can be conveniently plugged into any training process. As our future work, other applications of this loss will be explored.


{\small
\bibliographystyle{ieee}
\bibliography{refs}
}

\end{document}